\documentclass[doublecolumn,10pt] {IEEEtran}
\usepackage{cite}
\usepackage{amsmath,amssymb,amsfonts}
\usepackage{graphicx,color}
\usepackage{textcomp}
\usepackage{xcolor} 
\definecolor{elsevierblue}{HTML}{2fb2e2} 

\usepackage[colorlinks=true,
            linkcolor=elsevierblue,  
            citecolor=elsevierblue,  
            urlcolor=elsevierblue    
           ]{hyperref}
\usepackage{enumitem}
\usepackage{multirow}
\usepackage{titlesec}
\usepackage{academicons}
\usepackage{xcolor}
\newcommand*\emptycirc[1][1ex]{\tikz\draw (0,0) circle (#1);} 
\newcommand*\halfcirc[1][1ex]{%
  \begin{tikzpicture}
  \draw[fill] (0,0)-- (90:#1) arc (90:270:#1) -- cycle ;
  \draw (0,0) circle (#1);
  \end{tikzpicture}}
\newcommand*\fullcirc[1][1ex]{\tikz\fill (0,0) circle (#1);} 
\usepackage{subcaption} 
\usepackage{placeins}
\usepackage{booktabs}
\usepackage{tikz}
\usetikzlibrary{arrows.meta, positioning}
\usetikzlibrary{mindmap, shadows}
\usepackage{tabularx}
\usepackage{dblfloatfix} 

\usepackage[linesnumbered,ruled,vlined]{algorithm2e}

\definecolor{orcidlogocol}{HTML}{A6CE39}

\def\BibTeX{{\rm B\kern-.05em{\sc i\kern-.025em b}\kern-.08em
    T\kern-.1667em\lower.7ex\hbox{E}\kern-.125emX}}

\makeatletter
\renewenvironment{abstract}{%
  \begin{center}%
    {\normalfont\bfseries Abstract}%
  \end{center}%
  \normalfont 
}{%
  \par\vspace{0.5em}%
}

\let\oldIEEEkeywords\IEEEkeywords
\renewenvironment{IEEEkeywords}{%
  \oldIEEEkeywords
  \normalfont 
}{%
  \endlist
}
\makeatother

\begin{document}

\title{

FreeGNN: Continual Source--Free Graph Neural Network Adaptation for Renewable Energy Forecasting}

\author{%

\textbf{Abderaouf Bahi\textsuperscript{1\S}}~\href{https://orcid.org/0009-0003-7116-5080}{\includegraphics[height=1em]{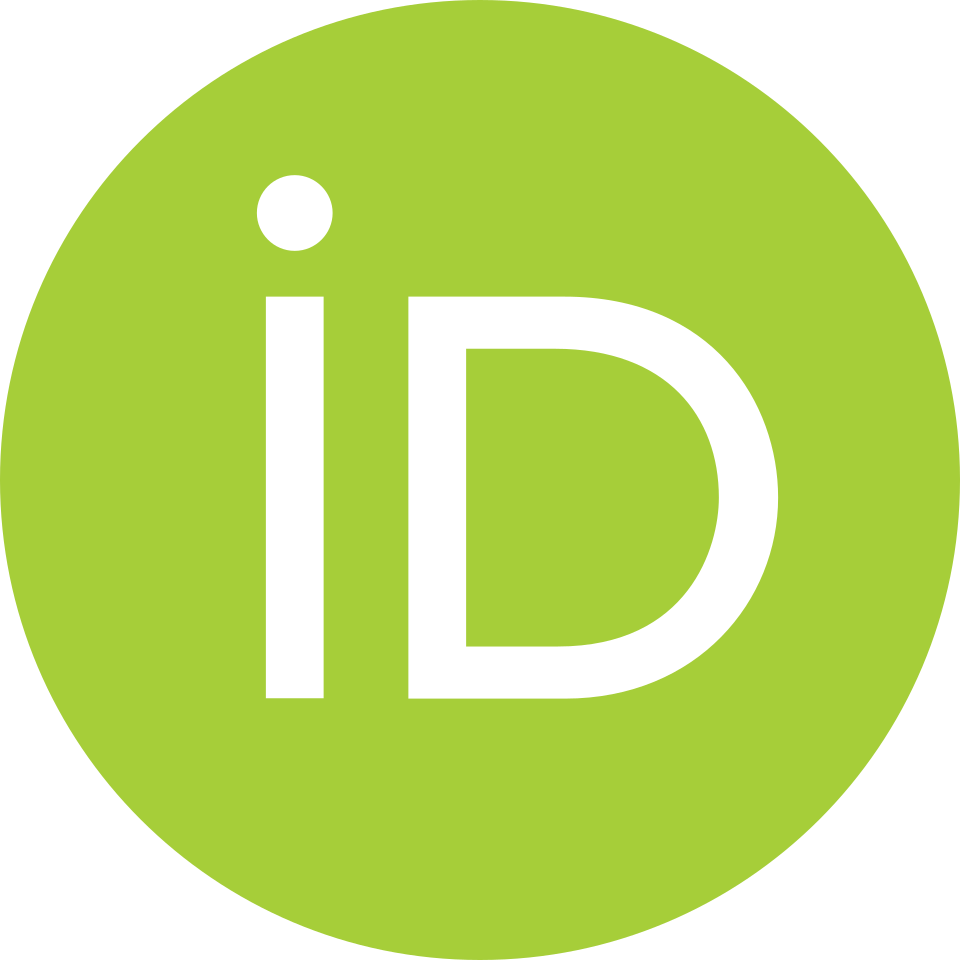}}, 
\textbf{Amel Ourici\textsuperscript{2}}~\href{https://orcid.org/0000-0001-5701-8576}{\includegraphics[height=1em]{ORCID_iD.png}}, 
\textbf{Ibtissem Gasmi\textsuperscript{1}}~\href{https://orcid.org/0000-0002-8939-1727}{\includegraphics[height=1em]{ORCID_iD.png}}, 
\textbf{Aida Derrablia\textsuperscript{1*}}~\href{https://orcid.org/0009-0005-3816-6297}{\includegraphics[height=1em]{ORCID_iD.png}}, 
\textbf{Warda Deghmane\textsuperscript{1*}}~\href{https://orcid.org/0009-0007-0362-1514}{\includegraphics[height=1em]{ORCID_iD.png}} and
\textbf{Mohamed~Amine~Ferrag\textsuperscript{3}}~\href{https://orcid.org/0000-0002-0632-3172}{\includegraphics[height=1em]{ORCID_iD.png}} \\[0.5em]
\textsuperscript{1}\small
Computer Science and Applied Mathematics Laboratory (LIMA),
Faculty of Science and Technology, Chadli Bendjedid University, 
P.O. Box 73, El Tarf 36000, Algeria \\[0.5em]
\textsuperscript{2}\small
Mathematical Modeling and Numerical Simulation Laboratory (LAM2SIN),
Faculty of Technology, Badji Mokhtar University, 
P.O. Box 12, Annaba 23000, Algeria \\[0.5em]
\textsuperscript{3}\small
College of Information Technology,
United Arab Emirates University, Al Ain, United Arab Emirates\\[0.5em]
\textsuperscript{\S}Corresponding author: \textbf{Abderaouf Bahi} (a.bahi@univ-eltarf.dz) \\[0.5em]
\textsuperscript{*}These authors contributed equally to this work.
}

\maketitle
\begin{abstract}
Accurate forecasting of renewable energy generation is essential for efficient grid management and sustainable power planning. However, traditional supervised models often require access to labeled data from the target site, which may be unavailable due to privacy, cost, or logistical constraints. In this work, we propose FreeGNN, a Continual Source-Free Graph Domain Adaptation framework that enables adaptive forecasting on unseen renewable energy sites without requiring source data or target labels. Our approach integrates a spatio-temporal Graph Neural Network (GNN) backbone with a teacher–student strategy, a memory replay mechanism to mitigate catastrophic forgetting, graph-based regularization to preserve spatial correlations, and a drift-aware weighting scheme to dynamically adjust adaptation strength during streaming updates. This combination allows the model to continuously adapt to non-stationary environmental conditions while maintaining robustness and stability.
We conduct extensive experiments on three real-world datasets: GEFCom2012, Solar PV, and Wind SCADA, encompassing multiple sites, temporal resolutions, and meteorological features. The ablation study confirms that each component—memory, graph regularization, drift-aware adaptation, and teacher–student strategy—contributes significantly to overall performance.
The experiments show that FreeGNN achieves an MAE of 5.237 and an RMSE of 7.123 on the GEFCom dataset, an MAE of 1.107 and an RMSE of 1.512 on the Solar PV dataset, and an MAE of 0.382 and an RMSE of 0.523 on the Wind SCADA dataset. These results demonstrate its ability to achieve accurate and robust forecasts in a source-free, continual learning setting, highlighting its potential for real-world deployment in adaptive renewable energy systems. For reproducibility, implementation details are available at \url{https://github.com/AraoufBh/FreeGNN}.
\end{abstract}



\def\BibTeX{{\rm B\kern-.05em{\sc i\kern-.025em b}\kern-.08em
    T\kern-.1667em\lower.7ex\hbox{E}\kern-.125emX}}

\section{Introduction}
\label{sec:intro}

The ability to accurately forecast renewable energy generation is crucial for maintaining efficiency and reliability in today’s power systems\cite{Sweeney2020Forecasting,Benti2023RenewableML,Bahi2025DroneRL}, enabling reliable grid operation, energy trading, and efficient integration of intermittent sources such as photovoltaic (PV) and wind power \cite{Das2018PVReview,Ackermann2012WindPower}.
In practice, forecasting models are deployed across geographically distributed sites with heterogeneous climates, sensor configurations, and operational conditions \cite{Mansoor2023HybridWindPV}.
As a result, a model trained on historical data from a subset of sites often suffers severe performance degradation when deployed on a new site or under new weather regimes \cite{Olczak2023PVDegradation}.
This distribution mismatch is further exacerbated by the nonstationarity of renewable energy streams \cite{Hammami2020OnlineLearning,datacenter}, where seasonal cycles, sensor recalibrations, equipment aging, and extreme weather events continuously alter the underlying data distribution \cite{Verma2024EnvironmentalMonitoring}.

Domain adaptation (DA) has emerged as a principled solution to mitigate distribution shifts by aligning source and target domains \cite{Zhang2026MuDiSGDA,Chang2026CUMDAN}.
However, most DA methods rely on the availability of source data during adaptation and assume that the target distribution is static.
These assumptions are rarely satisfied in operational renewable energy settings.
First, source data are frequently inaccessible at deployment time due to privacy constraints, ownership policies, and storage limitations \cite{Tang2016CloudSecurity}.
Second, target data arrive as a stream, and the target distribution evolves over time, making offline adaptation insufficient \cite{Cardellini2022StreamAdaptation}.
Consequently, renewable energy forecasting requires adaptation methods that operate continuously, using only unlabeled target data, and remain stable under drift.

A second key aspect of renewable energy forecasting is the strong spatial dependencies among sites \cite{Iung2023VREModeling,Park2018SpatialPrediction}.
Nearby PV plants are driven by correlated irradiance patterns \cite{Bao2025SolarIrradiance}, and wind farms within the same meteorological region share similar wind dynamics \cite{Millstein2019WindVariability}.
Spatio-temporal Graph Neural Networks (GNNs) \cite{Sahili2023STGNNsurvey} have recently demonstrated strong performance for forecasting tasks by explicitly modeling such relational structure.
Nevertheless, existing GNN-based forecasting models are typically trained offline and assume that the deployment environment matches the training distribution \cite{Zhang2023GNNCarbon}.
When applied to new sites or evolving climates, these models often fail, and naively fine-tuning them online can lead to unstable updates and catastrophic forgetting.

In this work, we study a realistic yet under-explored setting: Continual Source-Free Domain Adaptation (CSFDA) for graph-structured renewable energy time-series streams.
In CSFDA, a model is pretrained on multiple labeled source domains, then deployed on a target domain where (i) source data are not accessible, (ii) target labels are unavailable, and (iii) the model must adapt online as new target samples arrive.
This setting captures real operational constraints in renewable energy systems and introduces unique challenges beyond standard unsupervised domain adaptation or test-time adaptation.

To address these challenges, we propose FreeGNN, a continual source-free graph domain adaptation framework for renewable energy forecasting.
FreeGNN combines a spatio-temporal GNN forecasting backbone with an online teacher--student adaptation mechanism, graph-aware representation consistency, and a drift-aware memory replay strategy.
The teacher model provides stable pseudo-labels and consistency targets, while memory replay prevents catastrophic forgetting and improves robustness to recurring seasonal patterns.
Moreover, drift-aware weighting modulates adaptation strength over time, enabling stable updates during stationary periods and rapid adaptation under distribution shifts. We summarize our main contributions as follows:

\begin{itemize}
    \item We introduce a novel continual source-free domain adaptation framework that enables forecasting on unseen renewable energy sites without access to source data or target labels.  
    \item We propose a spatio-temporal GNN backbone combined with a teacher–student strategy, memory replay, graph regularization, and drift-aware adaptation, enabling stable and robust online learning under non-stationary conditions.  
    \item We perform extensive experiments on three real-world renewable energy datasets, demonstrating that each component significantly contributes to forecasting accuracy and achieves state-of-the-art results on evaluation metrics.  
    \item Our framework provides a practical solution for real-world deployment, enabling adaptive, continual forecasting across multiple sites while maintaining robustness to distribution shifts.
\end{itemize}

The remainder of the paper is organized as follows.
Section~\ref{sec:related} reviews related work.
Section~\ref{sec:prelim} introduces preliminaries and formally defines the problem.
Section~\ref{sec:method} presents the proposed FreeGNN method.
Section~\ref{sec:exp} reports experimental results.
Finally, Section~\ref{sec:concl} concludes the paper and discusses future directions.

\section{Related Work}
\label{sec:related}

Our work intersects with several lines of research, including domain adaptation for time series, source-free domain adaptation, graph domain adaptation, continual adaptation, and forecasting in renewable energy contexts. We review each area with emphasis on recent advances and their limitations relative to our proposed setting.

\subsection{Domain Adaptation for Time Series}

Time-series data pose unique challenges for domain adaptation due to temporal dependencies and non-stationarity. Zhang et al. (2024) propose a continual test-time domain adaptation framework for online machinery fault diagnosis, addressing continual covariate and label shifts via a teacher-student mechanism and class-balanced sampling \cite{tian2024continual}. Luo et al. (2026) explore adaptive virtual adversarial domain adaptation for time-series classification, aligning the source and target domains while preserving local smoothness in predictions \cite{lekshmi2026vasad}. Similarly, Jia et al. (2025) introduce a contrastive representation domain adaptation method for industrial cross-domain time series prediction to capture intra-domain relationships and mutual information \cite{jia2025contrastive}.

\subsection{Source-Free Domain Adaptation}

Source-free domain adaptation (SFDA) aims to adapt models to target domains without access to source data. Zhong et al. (2025) propose LEMON, a wavelet-based multi-scale temporal imputation method for time-series SFDA, that learns scale-specific domain-invariant representations \cite{zhong2025lemon}. Furqon et al. (2025) introduce TFDA, exploiting time-frequency synergy and contrastive learning to improve performance under SFTSDA settings \cite{furqon2025tfda}. Paeedeh et al. (2024) address cross-domain few-shot learning with adaptive transformer networks \cite{paeedeh2024adapter}. Ragab et al. (2026) present Evidential-MAPU, which leverages uncertainty estimation for source-free time-series adaptation \cite{ragab2026emapu}. Zhang et al. (2026) propose a distributed federated learning approach for source-free cross-domain remaining useful life prediction using graph convolutional neural networks \cite{zhang2026rul}.

\subsection{Graph and Source-Free Graph Domain Adaptation}

Graph-structured data introduces additional complexity in SFDA. Luo et al. (2024) propose GALA, a graph diffusion-based alignment with jigsaw method for source-free graph adaptation \cite{luo2024gala}. Zhang et al. (2024) introduce GraphCTA, a collaborative model-graph adaptation framework for source-free graph domain adaptation \cite{zhang2024graphcta}. Mao et al. (2024) develop SOGA for unsupervised source-free graph domain adaptation, preserving structural consistency on target graphs \cite{mao2024soga}. Luo et al. (2023) propose source-free progressive graph learning (SF-PGL) for open-set graph domain adaptation \cite{luo2023sfpgl}. Zhou et al. (2025) present CL-GNN, a continual learning GNN for dynamic wireless resource allocation \cite{zhou2025clgnn}.

\subsection{Continual and Online Adaptation}

Continual adaptation addresses streaming data scenarios with evolving distributions. Tian et al. (2024) introduce a continual test-time domain adaptation (CTDA) framework for online machinery fault diagnosis under dynamic operating conditions \cite{tian2024continual}. Yang et al. (2025) propose CSFADA, a continual source-free active domain adaptation framework for medical image segmentation across multiple centers \cite{yang2025csfada}. These methods combine teacher-student learning, domain reference selection, and distillation to mitigate catastrophic forgetting while adapting to evolving target distributions.

To summarize, existing works contribute valuable insights across DA, SFDA, graph adaptation, and time-series modeling. Yet, they exhibit the following limitations relative to our setting:
\begin{itemize}
    \item Most SFDA works focus on classification rather than forecasting.
    \item Graph SFDA methods address non-temporal graph tasks and lack mechanisms for time-series dynamics.
    \item Continual adaptation under source-free constraints for streaming time-series forecasting is underexplored.
    \item Application to renewable energy forecasting remains largely unexplored in a source-free, continual setup.
\end{itemize}
Our work aims to fill these gaps by proposing a continual source-free graph domain adaptation framework tailored for renewable energy time-series forecasting.

\begin{table*}[h!]
\centering
\caption{State of the art summary}
\label{tab:related_summary_colored}
\renewcommand{\arraystretch}{1.2}
\begin{tabular}{p{2.2cm} p{0.8cm} p{4cm} p{3.2cm}p{1.5cm}p{1.5cm}p{0.9cm}p{0.9cm}}
\toprule
\textbf{Reference} & \textbf{Year} & \textbf{Approach} & \textbf{Dataset} & 
\textbf{Continual Learning} & 
\textbf{Domain Adaptation} & 
\textbf{Source Free} & 
\textbf{Time Series} \\
\midrule
Zhang et al. \cite{zhang2026rul} & 2026 & Distributed SF RUL prediction  & NASA aircraft engines & \textcolor{red}{\emptycirc} & \textcolor{green}{\fullcirc} & \textcolor{green}{\fullcirc} & \textcolor{green}{\fullcirc} \\
Lekshmi et al. \cite{lekshmi2026vasad} & 2026 & VASAD: virtual adversarial TSC & Fault diagnosis & \textcolor{red}{\emptycirc} & \textcolor{green}{\fullcirc} & \textcolor{green}{\fullcirc} & \textcolor{green}{\fullcirc} \\
Ragab et al. \cite{ragab2026emapu} & 2026 & E-MAPU SF temporal imputation & Real-world time series & \textcolor{red}{\emptycirc} & \textcolor{green}{\fullcirc} & \textcolor{green}{\fullcirc} & \textcolor{green}{\fullcirc} \\
Zia et al. \cite{jia2025contrastive} & 2025 & Contrastive domain adaptation & CMAPSS & \textcolor{red}{\emptycirc} & \textcolor{green}{\fullcirc} & \textcolor{orange}{\halfcirc} & \textcolor{green}{\fullcirc} \\
Yang et al. \cite{yang2025csfada} & 2025 & CSFADA for tumor segmentation & Multi-center NPC images & \textcolor{green}{\fullcirc} & \textcolor{green}{\fullcirc} & \textcolor{green}{\fullcirc} & \textcolor{red}{\emptycirc} \\
Zhong et al. \cite{zhong2025lemon} & 2025 & Wavelet-based SFDA (LEMON) & Time series datasets & \textcolor{red}{\emptycirc} & \textcolor{green}{\fullcirc} & \textcolor{green}{\fullcirc} & \textcolor{green}{\fullcirc} \\
Furqon et al. \cite{furqon2025tfda} & 2025 & TFDA time-frequency SFDA & Time series datasets & \textcolor{red}{\emptycirc} & \textcolor{green}{\fullcirc} & \textcolor{green}{\fullcirc} & \textcolor{green}{\fullcirc} \\
Zhou et al. \cite{zhou2025clgnn} & 2025 & CL-GNN for wireless networks & Dynamic wireless network  & \textcolor{green}{\fullcirc} & \textcolor{green}{\fullcirc} & \textcolor{red}{\emptycirc} & \textcolor{red}{\emptycirc} \\
Paeedeh et al. \cite{paeedeh2024adapter} & 2024 & Adaptive transformer for  few-shot & BSCD-FSL benchmarks & \textcolor{red}{\emptycirc} & \textcolor{green}{\fullcirc} & \textcolor{orange}{\halfcirc} & \textcolor{red}{\emptycirc} \\
Zhang et al. \cite{zhang2024graphcta} & 2024 & GraphCTA collaborative SF-GDA & Public graph datasets & \textcolor{red}{\emptycirc} & \textcolor{green}{\fullcirc} & \textcolor{green}{\fullcirc} & \textcolor{red}{\emptycirc} \\
Mao et al. \cite{mao2024soga} & 2024 & SOGA SF graph domain adaptation & Cross-domain graph tasks & \textcolor{red}{\emptycirc} & \textcolor{green}{\fullcirc} & \textcolor{green}{\fullcirc} & \textcolor{red}{\emptycirc} \\
Tian et al. \cite{tian2024continual} & 2024 & CTDA with teacher-student & Machinery fault datasets & \textcolor{green}{\fullcirc} & \textcolor{green}{\fullcirc} & \textcolor{orange}{\halfcirc} & \textcolor{green}{\fullcirc} \\
Luo et al. \cite{luo2024gala} & 2024 & GALA: graph diffusion alignment & Graph benchmark datasets & \textcolor{red}{\emptycirc} & \textcolor{green}{\fullcirc} & \textcolor{green}{\fullcirc} & \textcolor{red}{\emptycirc} \\
Luo et al. \cite{luo2023sfpgl} & 2023 & SF-PGL for open-set graphs & Image/action recognition  & \textcolor{red}{\emptycirc} & \textcolor{green}{\fullcirc} & \textcolor{green}{\fullcirc} & \textcolor{red}{\emptycirc} \\
\midrule
This work  & 2026 & FreeGNN & Renewable energy datasets & \textcolor{green}{\fullcirc} & \textcolor{green}{\fullcirc} & \textcolor{green}{\fullcirc} & \textcolor{green}{\fullcirc} \\
\bottomrule
\end{tabular}
Not Considered (\textcolor{red}\emptycirc); Partial Consideration (\textcolor{orange}\halfcirc); Considered (\textcolor{green}\fullcirc);
\end{table*}

\section{Preliminaries and Problem Formulation}
\label{sec:prelim}

\subsection{Graph-Structured Renewable Energy Streams}
We consider a renewable energy system composed of multiple geographically distributed sites.
We model the system as a weighted graph $G=(V,E)$ with $|V|=N$ nodes.
Each node $v\in V$ corresponds to a site, and each edge $(u,v)\in E$ encodes a spatial or statistical dependency (e.g., geographical proximity or historical correlation).

Let $A \in \mathbb{R}^{N\times N}$ denote the weighted adjacency matrix of $G$.
Edge weights can be defined by distance-based kernels or correlation scores.
For example, a common choice is defined in Equation~\eqref{eq:adj_dist}.
\begin{equation}
A_{uv} = \exp\left(-\frac{\text{dist}(u,v)^2}{\kappa}\right)\cdot \mathbb{I}\left(\text{dist}(u,v)\le r\right),
\label{eq:adj_dist}
\end{equation}
where $\kappa$ controls decay and $r$ is a cutoff radius.
Alternatively, one may define $A_{uv}$ using Pearson correlation between historical power profiles.

At each time step $t$, node $v$ is associated with a feature vector $x_t^v \in \mathbb{R}^{d}$.
Typical features include past power generation, irradiance, temperature, wind speed, humidity, and calendar variables.
We denote the graph signal at time $t$ as defined in Equation~\eqref{eq:grph-sign}.
\begin{equation}
X_t = [x_t^1, \dots, x_t^N]^\top \in \mathbb{R}^{N\times d}.
\label{eq:grph-sign}
\end{equation}
The target variable is the renewable power output $y_t^v \in \mathbb{R}$ for each node, collected as defined in Equation~\eqref{eq:renpow-out}.
\begin{equation}
Y_t = [y_t^1, \dots, y_t^N]^\top \in \mathbb{R}^{N}.
\label{eq:renpow-out}
\end{equation}

\subsection{Forecasting Task}
\label{subsec:forecast_task}

We study rolling-window multi-horizon forecasting.
Given a window length $w$ and forecasting horizon $h$, the model receives the past sequence defined in Equation~\eqref{eq:past-seq}.
\begin{equation}
\mathbf{X}_{t-w+1:t} = \{X_{t-w+1}, \dots, X_t\},
\label{eq:past-seq}
\end{equation}
and predicts the future renewable generation as defined in Equation~\eqref{eq:forecast}.
\begin{equation}
\hat{Y}_{t+h} = f_\theta(\mathbf{X}_{t-w+1:t}, G),
\label{eq:forecast}
\end{equation}
where $f_\theta$ is a spatio-temporal forecasting model (e.g., GNN-based) with parameters $\theta$.

In supervised settings, $f_\theta$ is trained by minimizing a forecasting loss such as MAE defined in Equation~\eqref{eq:fore-loss}.
\begin{equation}
\ell(\hat{Y}_{t+h}, Y_{t+h}) = \|\hat{Y}_{t+h}-Y_{t+h}\|_1.
\label{eq:fore-loss}
\end{equation}

\subsection{Domains and Distribution Shifts}
\label{subsec:domains}

We formalize domain shift in renewable energy forecasting by accounting for cross-site and cross-region variations.
A domain corresponds to a data-generating process characterized by a joint distribution over time-series windows and labels as defined in Equation~\eqref{eq:doma-la}.
\begin{equation}
\mathcal{D} : \ (\mathbf{X}_{t-w+1:t}, Y_{t+h}) \sim P(\mathbf{X},Y).
\label{eq:doma-la}
\end{equation}

We assume access to $K$ labeled \emph{source} domains as defined in Equation~\eqref{eq:ksource}.
\begin{equation}
\{\mathcal{D}_S^k\}_{k=1}^K,
\label{eq:ksource}
\end{equation}
and one \emph{target} domain $\mathcal{D}_T$.
The domain shift is reflected using Equation~\eqref{eq:dom-shift}.
\begin{equation}
P_S(\mathbf{X},Y) \neq P_T(\mathbf{X},Y),
\label{eq:dom-shift}
\end{equation}
where $P_S$ denotes the mixture of source distributions.

In practice, shifts arise due to geographic and climate differences, seasonal patterns, sensor calibration and noise, and local operational conditions.
Unlike standard domain adaptation, the target distribution is non-stationary over time as defined in Equation~\eqref{eq:nonstationary}.
\begin{equation}
P_T^t(\mathbf{X},Y) \neq P_T^{t+\Delta}(\mathbf{X},Y),
\label{eq:nonstationary}
\end{equation}
capturing gradual drift and abrupt changes (e.g., extreme weather events).

\subsection{Continual Source-Free Domain Adaptation Setting}
\label{subsec:csfda_setting}

We study continual source-free domain adaptation for forecasting.
The learning process is divided into two phases:

\paragraph{(1) Multi-source pretraining.}
During pretraining, the learner has access to labeled source data from $\{\mathcal{D}_S^k\}_{k=1}^K$.
This phase produces an initial model $f_{\theta_0}$.

\paragraph{(2) Target deployment and online adaptation.}
After deployment, the learner observes only an unlabeled target stream as defined in Equation~\eqref{eq:targ-stream}.
\begin{equation}
\mathcal{S}_T = \{X_t^T\}_{t=1}^\infty.
\label{eq:targ-stream}
\end{equation}
At adaptation time:
\begin{itemize}
    \item No source data are accessible 
    \item No target labels are available during online adaptation.
    \item The model must update continually as new target samples arrive.
\end{itemize}

We emphasize that the source-free constraint is stricter than standard unsupervised domain adaptation, as it forbids revisiting source samples or computing source statistics during deployment.

\subsection{Objective}
\label{subsec:objective_prelim}

Our goal is to learn a sequence of models $\{f_{\theta_t}\}_{t\ge 0}$ such that forecasting performance on the evolving target stream is minimized over time.
Let $\theta_t$ denote the parameters after observing target data up to time $t$.
We aim to minimize the online risk as defined in Equation~\eqref{eq:online_risk}.
\begin{equation}
\min_{\{\theta_t\}}
\sum_{t \ge w}
\mathbb{E}\left[
\ell\left(
f_{\theta_t}(\mathbf{X}_{t-w+1:t}^T,G),
Y_{t+h}^T
\right)
\right],
\label{eq:online_risk}
\end{equation}
subject to the constraint that $Y_{t+h}^T$ is never used during adaptation. In other words, labels are only used for evaluation, not for updating the deployed model.

\section{Method}
\label{sec:method}

This section details the FreeGNN model. Given a model pretrained on multiple labeled source domains, we deploy it on a target domain where only an unlabeled stream is observed and source data are inaccessible.
Our goal is to continuously adapt the model online while preventing catastrophic forgetting and ensuring stable performance under non-stationary drift.

\begin{figure*}[h!]
    \centering    \includegraphics[width=\linewidth,height=0.5\textheight,keepaspectratio]{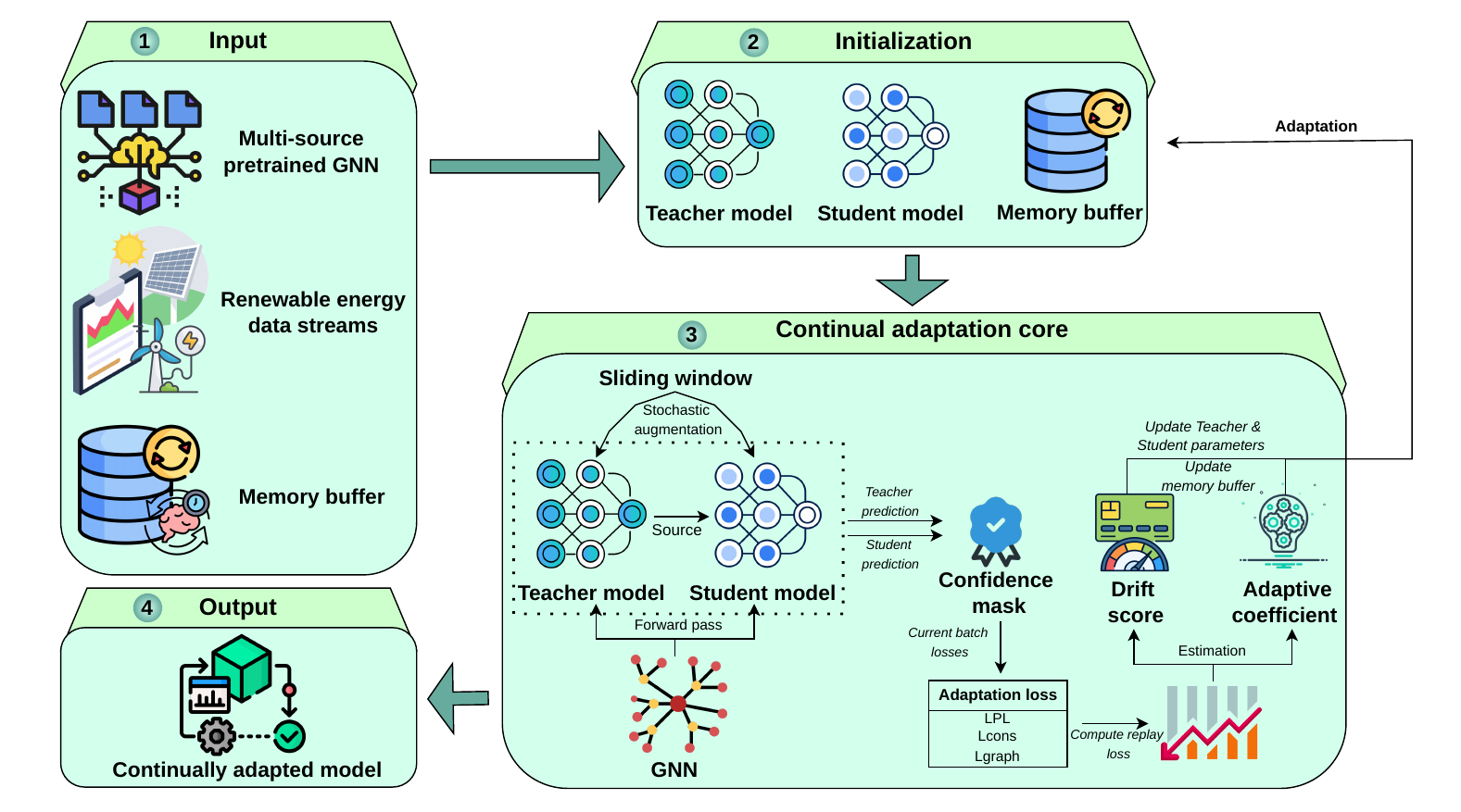}
    \caption{FreeGNN's General workflow}
    \label{fig:genworkflow}
\end{figure*}

Figure~\ref{fig:genworkflow} illustrates the proposed framework, which is summarized in Algorithm~\ref{alg:csfgda}.
Our method consists of four main components:
(i) a spatio-temporal GNN forecasting backbone,
(ii) a source pretraining stage,
(iii) a source-free online adaptation stage using teacher-student consistency and entropy-based regularization,
and (iv) a drift-aware memory replay mechanism to stabilize learning and mitigate forgetting.

\subsection{Data Preprocessing}
\label{subsec:data_preprocessing}

Prior to the implementation of FreGNN, careful preprocessing of the datasets was performed to ensure data quality, consistency, and suitability for temporal forecasting.

Extensive datasets from multiple renewable energy sites may contain missing or duplicate entries. Samples with missing essential data, such as power measurements or meteorological variables, were removed. Duplicate records were eliminated to guarantee that each timestamped observation is unique.

For numerical variables, missing values were imputed using the mean of observed values as defined in Equation~\eqref{eq:misv}.
\begin{equation}
x'_i = \frac{1}{n} \sum_{i=1}^{n} x_i
\label{eq:misv}
\end{equation}
For categorical variables, such as weather type or operational status, missing entries were replaced with the mode of the observed category as defined in Equation~\eqref{eq:mod}.
\begin{equation}
x'_i = \text{Mode}(x)
\label{eq:mod}
\end{equation}
This strategy preserves data consistency while minimizing the introduction of bias.

Categorical features, including site type, region, or operational flags, were converted into a machine-readable format using one-hot encoding. For instance, a categorical variable with three possible values $\{A, B, C\}$ is represented as defined in Equation~\eqref{eq:cas}.
\begin{equation}
Y = 
\begin{cases}
(1, 0, 0) & \text{if } y = A \\
(0, 1, 0) & \text{if } y = B \\
(0, 0, 1) & \text{if } y = C
\end{cases}
\label{eq:cas}
\end{equation}
This transformation prevents the introduction of unintended ordinal relationships among categories.
 
Timestamps were normalized to capture the temporal evolution of power generation and meteorological influences, preventing older measurements from disproportionately affecting the model. Each timestamp $t_i$ was scaled as defined in Equation~\eqref{eq:tmsta}.
\begin{equation}
t'_i = \frac{t_i - \min(T)}{\max(T) - \min(T)}
\label{eq:tmsta}
\end{equation}
where $T$ is the set of all timestamps in the dataset.

The preprocessed datasets were then split into training and testing sets, with 80\% of the data used for training and 20\% for testing. This split ensures that the model has sufficient data to learn robust temporal and spatial patterns while reserving a representative portion of unseen data to reliably evaluate forecasting performance and generalization under realistic streaming conditions.

\subsection{Spatio-Temporal Graph Forecasting Backbone}
\label{subsec:backbone}

Let $G=(V,E)$ be a graph with $|V|=N$ nodes.
Each node $v \in V$ represents a renewable energy site.
Edges encode spatial proximity or historical correlation.
At time $t$, each node is associated with an input feature vector $x_t^v \in \mathbb{R}^{d}$.
We denote the multivariate graph signal as $X_t = [x_t^1, \dots, x_t^N]^\top \in \mathbb{R}^{N \times d}$.

We consider a rolling-window forecasting setting. 
Given a historical window of length $w$, the model receives the input sequence as defined in Equation~\eqref{eq:window}.

\begin{equation}
\mathbf{X}_{t-w+1:t} = \{ X_{t-w+1}, \dots, X_t \}
\label{eq:window}
\end{equation}

where $\mathbf{X}_{t-w+1:t}$ denotes the multivariate observations collected from time step $t-w+1$ to $t$.

The model then predicts the renewable power generation at forecasting horizon $h$ as defined in Equation~\eqref{eq:forecast}.

\begin{equation}
\hat{Y}_{t+h} = f_\theta(\mathbf{X}_{t-w+1:t}, G) \in \mathbb{R}^{N}
\label{eq:forecast}
\end{equation}

where $f_\theta$ represents a spatio-temporal Graph Neural Network (GNN) parameterized by $\theta$, $G$ denotes the underlying graph structure encoding spatial dependencies, $N$ is the number of nodes (e.g., wind turbines or solar stations), and $\hat{Y}_{t+h}$ is the predicted power vector at time step $t+h$.

We first map each node's historical sequence to a latent embedding using a temporal encoder $T_\theta(\cdot)$ as defined in Equation~\eqref{eq:tempo}.
\begin{equation}
H_t = T_\theta(\mathbf{X}_{t-w+1:t}) \in \mathbb{R}^{N \times d_h},
\label{eq:tempo}
\end{equation}
where $H_t[v]$ summarizes the temporal context for node $v$.

We then perform graph propagation to capture cross-site dependencies.
Let $\mathcal{N}(v)$ denote neighbors of $v$.
A generic message passing layer is written as defined in Equation~\eqref{eq:messpass}.
\begin{equation}
\tilde{H}_t[v] = \phi\left(
H_t[v],\ 
\textstyle \sum_{u \in \mathcal{N}(v)} \alpha_{vu}\, \psi(H_t[u])
\right),
\label{eq:messpass}
\end{equation}
where $\psi(\cdot)$ and $\phi(\cdot)$ are learnable functions, and $\alpha_{vu}$ are normalized edge weights derived from adjacency $A$.

Finally, a prediction head $g_\theta(\cdot)$ produces the forecast as defined in Equation~\eqref{eq:predhead}.
\begin{equation}
\hat{Y}_{t+h} = g_\theta(\tilde{H}_t).
\label{eq:predhead}
\end{equation}

\begin{figure}[h!]
    \centering
    \includegraphics[width=0.6\linewidth,height=\textheight,keepaspectratio]{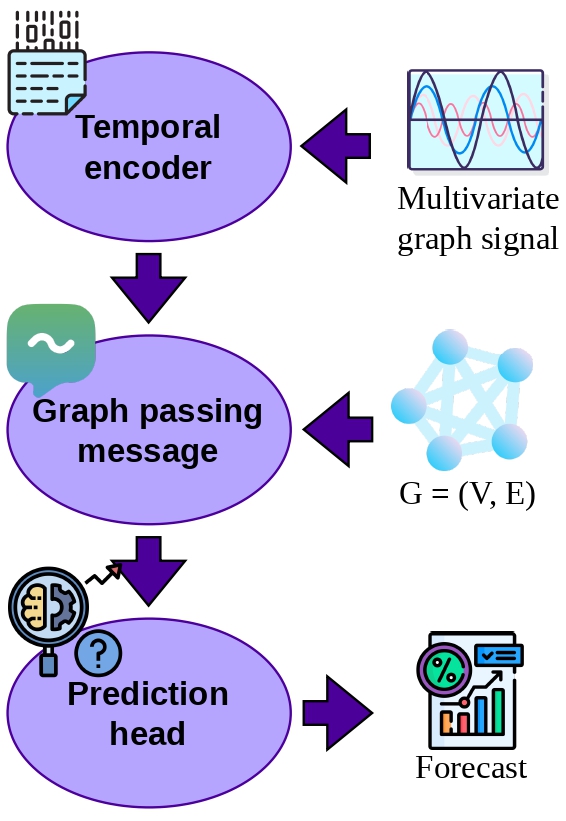}
    \caption{Spatio-temporal graph forecasting overall procedure}
    \label{fig:forcast}
\end{figure}

A simplified workflow of the overall procedure is presented in Figure ~\ref{fig:forcast}
\subsection{Multi-Source Supervised Pretraining}
\label{subsec:pretraining}

We assume access to labeled datasets from $K$ source domains $\{\mathcal{D}_S^k\}_{k=1}^K$.
Each source domain corresponds to a set of sites or a region with consistent distribution.
A sample from source domain $k$ is denoted as $(\mathbf{X}_{t-w+1:t}^{(k)}, Y_{t+h}^{(k)})$.

We pretrain the backbone $f_\theta$ by minimizing the supervised forecasting loss across sources as defined in Equation~\eqref{eq:lsup}.
\begin{equation}
\mathcal{L}_{\text{sup}}(\theta)
=
\sum_{k=1}^{K}
\mathbb{E}_{(\mathbf{X},Y)\sim \mathcal{D}_S^k}
\left[
\ell\left(f_\theta(\mathbf{X},G), Y\right)
\right],
\label{eq:lsup}
\end{equation}
where $\ell(\cdot,\cdot)$ is typically $\ell_1$ (MAE) or Huber loss for robustness.
This yields an initial model $\theta_0$.

In renewable energy systems, the target stream may exhibit substantial domain shift due to geography, climate, sensor calibration, and seasonal effects.
Thus, a source-trained model typically suffers significant degradation when deployed, motivating source-free online adaptation.

\subsection{Continual Source-Free Online Adaptation}
\label{subsec:csfda}

After deployment, we observe an unlabeled target stream $\{X_t^T\}_{t=1}^\infty$.
At each time step $t$, we must update the model using only the current (and previously seen) target data.
Importantly, no source data are available during adaptation.

We propose a teacher-student framework combined with graph-aware consistency and memory replay.
The student model $f_{\theta}$ is updated online.
The teacher model $f_{\bar{\theta}}$ is an exponential moving average (EMA) of the student as defined in Equation~\eqref{eq:ema}.
\begin{equation}
\bar{\theta} \leftarrow \mu \bar{\theta} + (1-\mu)\theta,
\label{eq:ema}
\end{equation}
where $\mu \in [0,1)$ controls stability.

EMA teachers are widely used to provide stable pseudo-labels under noisy updates.

Since the target stream is unlabeled, we generate pseudo-labels from the teacher.
Given an input window $\mathbf{X}_{t-w+1:t}^T$, the teacher do a prediction as defined in Equation~\eqref{eq:techpred}.
\begin{equation}
\tilde{Y}_{t+h}^T = f_{\bar{\theta}}(\mathbf{X}_{t-w+1:t}^T, G).
\label{eq:techpred}
\end{equation}
We then update the student to match these predictions as defined in Equation~\eqref{eq:lpl}.
\begin{equation}
\mathcal{L}_{\text{PL}}
=
\mathbb{E}_{\mathbf{X} \sim \mathcal{D}_T}
\left[
\ell\left(
f_\theta(\mathbf{X},G),
\tilde{Y}
\right)
\right].
\label{eq:lpl}
\end{equation}

A key challenge is that pseudo-labels can be unreliable, especially early in adaptation.
We therefore apply confidence masking.
For each node $v$, we define a confidence score based on the teacher's predictive stability under perturbations (described next).
Only nodes with confidence above threshold $\tau$ contribute to $\mathcal{L}_{\text{PL}}$.

Renewable energy time-series exhibit smooth dynamics and strong temporal continuity.
We enforce prediction invariance under stochastic time-series augmentations.
Let $\mathcal{A}(\cdot)$ denote an augmentation operator such as:
(i) jittering (small Gaussian noise),
(ii) time masking,
(iii) scaling,
(iv) slight time warping.

For an input window $\mathbf{X}$, we generate two views as defined in Equation~\eqref{eq:2vie}.
\begin{equation}
\mathbf{X}^{(1)} = \mathcal{A}_1(\mathbf{X}), \quad
\mathbf{X}^{(2)} = \mathcal{A}_2(\mathbf{X}).
\label{eq:2vie}
\end{equation}
The teacher produces a target prediction from view 1 as defined in Equation~\eqref{eq:vie1}.
\begin{equation}
\tilde{Y}^{(1)} = f_{\bar{\theta}}(\mathbf{X}^{(1)},G),
\label{eq:vie1}
\end{equation}
while the student predicts from view 2 as defined in Equation~\eqref{eq:vie2}.
\begin{equation}
\hat{Y}^{(2)} = f_{\theta}(\mathbf{X}^{(2)},G).
\label{eq:vie2}
\end{equation}
We minimize the consistency loss as defined in Equation~\eqref{eq:lcons}.
\begin{equation}
\mathcal{L}_{\text{cons}}
=
\mathbb{E}_{\mathbf{X}\sim \mathcal{D}_T}
\left[
\|\hat{Y}^{(2)} - \tilde{Y}^{(1)}\|_1
\right].
\label{eq:lcons}
\end{equation}

We compute the node-wise confidence score as defined in Equation~\eqref{eq:nodwis}.
\begin{equation}
c_v = \exp\left(
-\frac{|\tilde{Y}^{(1)}[v] - \tilde{Y}^{(2)}[v]|}{\sigma}
\right),
\label{eq:nodwis}
\end{equation}
where $\tilde{Y}^{(2)}=f_{\bar{\theta}}(\mathbf{X}^{(2)},G)$ and $\sigma$ is a temperature.
Nodes with $c_v \ge \tau$ are included in losses \eqref{eq:lpl}--\eqref{eq:lcons}.
This avoids reinforcing unreliable pseudo-labels.

A unique aspect of our setting is that renewable energy sites are not independent.
Nearby or correlated sites share weather patterns and seasonal dynamics.
Thus, adaptation should preserve relational structure across nodes.

We propose a graph-aware regularization on node embeddings.
Let $Z_t \in \mathbb{R}^{N \times d_z}$ denote intermediate node representations produced by the student.
We enforce that connected nodes have similar representations, weighted by edge strength as defined in Equation~\eqref{eq:lgraph}.
\begin{equation}
\mathcal{L}_{\text{graph}}
=
\mathbb{E}_{\mathbf{X}\sim \mathcal{D}_T}
\left[
\sum_{(u,v)\in E} w_{uv}\, \|Z_t[u] - Z_t[v]\|_2^2
\right],
\label{eq:lgraph}
\end{equation}
where $w_{uv}$ are normalized adjacency weights.

Directly smoothing predictions can oversmooth and degrade forecasting accuracy.
By regularizing embeddings, the model retains expressive prediction capacity while still leveraging cross-site structure.

To further encourage confident predictions on the target stream, we optionally minimize predictive entropy.
Since forecasting is continuous-valued, we adopt an uncertainty proxy based on teacher-student disagreement as defined in Equation~\eqref{eq:lent}.
\begin{equation}
\mathcal{L}_{\text{ent}}
=
\mathbb{E}_{\mathbf{X}\sim \mathcal{D}_T}
\left[
\|f_{\theta}(\mathbf{X},G) - f_{\bar{\theta}}(\mathbf{X},G)\|_2^2
\right].
\label{eq:lent}
\end{equation}
This term reduces stochasticity and stabilizes online updates.

\subsection{Drift-Aware Memory Replay}
\label{subsec:memory}

A central challenge in continual adaptation is catastrophic forgetting.
In non-stationary renewable energy streams, the target distribution evolves with seasons and weather regimes.
Naive online self-training may overfit to the most recent data and degrade performance on recurring patterns like seasonal cycles.

We introduce a small memory buffer $\mathcal{M}$ of size $B$ storing representative windows from the target stream.
At each adaptation step, we sample a mini-batch from the buffer and jointly optimize the student on current data and replayed samples.

We use a reservoir sampling strategy to maintain diversity in $\mathcal{M}$ under streaming constraints.
At time step $t$, the current window $\mathbf{X}_t$ is inserted into $\mathcal{M}$ with probability $\min(1, B/t)$.
If inserted, it replaces a uniformly sampled element.
This yields an unbiased approximation of the full stream distribution.

\subsubsection{Replay loss}
For replayed windows $\mathbf{X}\in \mathcal{M}$, we compute the same teacher-student consistency losses as defined in Equation~\eqref{eq:lreplay}.
\begin{equation}
\mathcal{L}_{\text{replay}}
=
\mathcal{L}_{\text{cons}}(\mathcal{M})
+
\mathcal{L}_{\text{graph}}(\mathcal{M}),
\label{eq:lreplay}
\end{equation}
where losses are evaluated over replayed samples.
This encourages stable representations and reduces drift-induced forgetting.

To explicitly account for non-stationary drift, we dynamically weight the adaptation loss using a drift score.
We estimate drift by measuring the discrepancy between current embeddings and memory embeddings as defined in Equation~\eqref{eq:drift}.
\begin{equation}
d_t =
\left\|
\frac{1}{N}\sum_{v=1}^{N} Z_t[v]
-
\frac{1}{|\mathcal{M}|}\sum_{\mathbf{X}\in \mathcal{M}}
\left(\frac{1}{N}\sum_{v=1}^{N} Z_{\mathbf{X}}[v]\right)
\right\|_2.
\label{eq:drift}
\end{equation}
We then define a drift-aware coefficient as presented in Equation~\eqref{eq:lambda_drift}.
\begin{equation}
\lambda_t = \text{sigmoid}\left(\gamma(d_t - \delta)\right),
\label{eq:lambda_drift}
\end{equation}
where $\delta$ is a drift threshold and $\gamma$ controls sharpness.
Intuitively, when drift is low, the model performs mild updates; when drift increases, the model adapts more aggressively.

The final online adaptation objective combines all components as defined in Equation~\eqref{eq:fullobj}.
\begin{equation}
\mathcal{L}_{\text{adapt}}
=
\lambda_t
\left(
\lambda_1 \mathcal{L}_{\text{PL}}
+
\lambda_2 \mathcal{L}_{\text{cons}}
\right)
+
\lambda_3 \mathcal{L}_{\text{graph}}
+
\lambda_4 \mathcal{L}_{\text{replay}}
+
\lambda_5 \mathcal{L}_{\text{ent}}.
\label{eq:fullobj}
\end{equation}
We update the student parameters with SGD/Adam as defined in Equation~\eqref{eq:sgd}.
\begin{equation}
\theta \leftarrow \theta - \eta \nabla_\theta \mathcal{L}_{\text{adapt}},
\label{eq:sgd}
\end{equation}
then update the teacher using EMA \eqref{eq:ema}.

In early deployment, pseudo-labels are noisy.
We therefore apply a warm-up stage where only $\mathcal{L}_{\text{cons}}$ and $\mathcal{L}_{\text{graph}}$ are used for the first $T_0$ steps.
Afterward, $\mathcal{L}_{\text{PL}}$ is enabled with confidence masking.

\subsection{Complexity and Deployment Considerations}
\label{subsec:complexity}

Let $N$ denote the number of nodes, $|E|$ the number of edges in the graph, $w$ the temporal window length, and $d_h$ the hidden feature dimension.  
The computational complexity of a single forward pass is given by Equation~\eqref{eq:complexity}.

\begin{equation}
\mathcal{O}(N \cdot w \cdot d_h) 
+ 
\mathcal{O}(|E| \cdot d_h)
\label{eq:complexity}
\end{equation}

where the first term $\mathcal{O}(N \cdot w \cdot d_h)$ corresponds to the temporal encoding over the sliding window, and the second term $\mathcal{O}(|E| \cdot d_h)$ accounts for graph message passing operations across edges.
During online adaptation, one (or a small number of) gradient update step(s) are performed per time step. Therefore, the overall computational overhead remains linear with respect to the stream length, making the proposed approach suitable for near real-time deployment.

The buffer stores $B$ windows of size $w \times N \times d$.
Thus memory cost is $\mathcal{O}(B w N d)$.
In practice, we use small $B$ to fit edge devices.

Teacher-student EMA and confidence masking reduce confirmation bias.
Replay prevents rapid forgetting.
Drift-aware weighting avoids overly aggressive updates when the stream is stable, improving robustness.

In summary, the proposed FreeGNN framework enables continual adaptation of spatio-temporal GNNs under realistic constraints: no source data, no target labels, and non-stationary streaming conditions.
\begin{algorithm}[t!]
\small
\caption{Pseudo-code of FreeGNN approach}
\label{alg:csfgda}

\textbf{Input:}\\
\hspace*{1em}Multi-source pretrained model parameters $\theta_0$,\\
\hspace*{1em}target data stream $\{X_t\}_{t\ge1}$, graph $G$,\\
\hspace*{1em}memory buffer size $B$, window size $w$.

\textbf{Output:}\\
\hspace*{1em}Adapted model parameters $\theta$.

\textbf{Step 1: Initialization:}\\
\hspace*{1em}\textbf{1.1:} Initialize student model $\theta \leftarrow \theta_0$.\\
\hspace*{1em}\textbf{1.2:} Initialize teacher model $\bar{\theta} \leftarrow \theta_0$.\\
\hspace*{1em}\textbf{1.3:} Initialize memory buffer $\mathcal{M} \leftarrow \emptyset$.

\textbf{Step 2: Continual Adaptation Loop:}\\
\hspace*{1em}\textbf{2.1:} \textbf{for} $t = w, w+1, \dots$ \textbf{do}

\hspace*{2em}\textbf{2.2:} Construct current sliding window $\mathbf{X}_{t-w+1:t}$.\\
\hspace*{2em}\textbf{2.3:} Sample replay batch $\mathbf{X}^r \sim \mathcal{M}$.\\
\hspace*{2em}\textbf{2.4:} Generate two stochastic augmentations $\mathbf{X}^{(1)}, \mathbf{X}^{(2)}$.\\

\hspace*{2em}\textbf{2.5:} Teacher prediction: $\tilde{Y}^{(1)} = f_{\bar{\theta}}(\mathbf{X}^{(1)}, G)$.\\
\hspace*{2em}\textbf{2.6:} Student prediction: $\hat{Y}^{(2)} = f_{\theta}(\mathbf{X}^{(2)}, G)$.\\

\hspace*{2em}\textbf{2.7:} Compute confidence mask based on teacher disagreement.\\
\hspace*{2em}\textbf{2.8:} Compute adaptation losses on current batch: 
$\mathcal{L}_{\text{PL}}, \mathcal{L}_{\text{cons}}, \mathcal{L}_{\text{graph}}$.\\
\hspace*{2em}\textbf{2.9:} Compute replay losses on $\mathbf{X}^r$.\\

\hspace*{2em}\textbf{2.10:} Estimate drift score $d_t$ and adaptive coefficient $\lambda_t$.\\
\hspace*{2em}\textbf{2.11:} Update student parameters:
$\theta \leftarrow \theta - \eta \nabla_\theta \mathcal{L}_{\text{adapt}}$.\\
\hspace*{2em}\textbf{2.12:} Update teacher parameters:
$\bar{\theta} \leftarrow \mu \bar{\theta} + (1-\mu)\theta$.\\
\hspace*{2em}\textbf{2.13:} Update memory buffer $\mathcal{M}$ using reservoir sampling.\\

\hspace*{1em}\textbf{2.14:} \textbf{end for}

\textbf{Step 3: Return Adapted Model.}

\end{algorithm}
\section{Experiments}
\label{sec:exp}

To thoroughly evaluate the proposed FreeGNN framework, we conduct extensive experiments and provide a detailed analysis of the results. Our study is designed to answer the following research questions:

\begin{itemize}
    \item \textbf{RQ1:} Does FreeGNN improve forecasting accuracy under source-free domain adaptation compared to non-adaptive and state-of-the-art baselines?

    \item \textbf{RQ2:} How does FreeGNN perform across different renewable energy domains?

    \item \textbf{RQ3:} What is the contribution of each component of FreeGNN to the overall performance?

    \item \textbf{RQ4:} How sensitive is FreeGNN to key hyperparameters?

    \item \textbf{RQ5:} What is the computational overhead of online adaptation in terms of runtime and memory consumption?
\end{itemize}

\subsection{Datasets}

We evaluate our framework on three real-world renewable energy benchmarks covering wind farms, photovoltaic plants, and turbine-level SCADA measurements. These datasets exhibit spatial variability, seasonal effects, and distribution shifts across sites, making them suitable for source-free domain adaptation in spatio-temporal forecasting.

\paragraph{GEFCom2012 -- Wind Forecasting.}
The wind forecasting track of the Global Energy Forecasting Competition 2012 (GEFCom2012)\footnote{\url{https://www.kaggle.com/competitions/global-energy-forecasting/data}} contains hourly wind power production from seven geographically distributed wind farms over approximately two years (2009--2010). Power outputs are normalized between 0 and 1. In addition, numerical weather prediction (NWP) forecasts are provided, including zonal and meridional wind components ($u$, $v$), wind speed ($ws$), and wind direction ($wd$). We treat each wind farm as a graph node and construct spatial edges based on pairwise correlation of historical production profiles.

\paragraph{Solar Power Generation Dataset.}
The photovoltaic dataset\footnote{\url{https://www.kaggle.com/datasets/anikannal/solar-power-generation-data}} contains inverter-level power generation and plant-level meteorological sensor readings from two solar plants in India, recorded at 15-minute resolution over 34 days. Variables include DC power, AC power, ambient temperature, and irradiance-related measurements. Each plant is considered a graph node, with inter-site connections defined via feature similarity.

\paragraph{Wind Turbine SCADA Dataset.}
The wind turbine SCADA dataset\footnote{\url{https://www.kaggle.com/datasets/berkerisen/wind-turbine-scada-dataset}} consists of 10-minute operational measurements from a utility-scale wind turbine over one year. Recorded variables include active power output, wind speed, theoretical power curve, and wind direction. This dataset is used to evaluate adaptation under fine-grained temporal resolution and operational drift.

Table~\ref{tab:datasets} summarizes the main characteristics of the datasets.

\begin{table*}[t]
\centering
\caption{Summary of renewable energy datasets used in our experiments}
\label{tab:datasets}
\begin{tabular}{lcccccc}
\toprule
Dataset & \#Sites & Samples & Duration & Freq. & Power Vars & Met. Vars  \\
\midrule
GEFCom2012 
& 7 farms 
& $\sim$17,500/site 
& 2 years 
& 1h 
& 1 
& 4 (u, v, ws, wd)  \\

Solar PV  
& 2 plants 
& $\sim$3,200/plant 
& 34 days 
& 15 min 
& 2 (DC, AC) 
& 4--6  \\

Wind SCADA 
& 1 turbine 
& $\sim$52,000 
& 1 year 
& 10 min 
& 1 
& 2 (speed, direction)  \\

\bottomrule
\end{tabular}
\end{table*}

\subsection{Evaluation protocol}
To ensure a realistic evaluation under streaming conditions, we adopt a pretrain--deploy--adapt protocol to evaluate the effectiveness of the proposed approach:

\begin{enumerate}
    \item Train $f_{\theta_0}$ on labeled multi-source data.
    \item Initialize the target model with $\theta_0$.
    \item Update the model online using only the unlabeled target stream.
    \item Measure forecasting accuracy at each time step or over rolling windows.
\end{enumerate}

To quantitatively assess forecasting performance, we report the following evaluation metrics:

\paragraph{Mean Absolute Error (MAE).}
\begin{align}
\text{MAE} = \frac{1}{N}\sum_{i=1}^{N} \left| y_i - \hat{y}_i \right|
\label{eq:mae}
\end{align}
where $N$ denotes the number of samples, $y_i$ is the ground-truth value, and $\hat{y}_i$ is the predicted value at time step $i$, as defined in Equation~\eqref{eq:mae}.

\paragraph{Root Mean Squared Error (RMSE).}
\begin{align}
\text{RMSE} = \sqrt{\frac{1}{N}\sum_{i=1}^{N} \left( y_i - \hat{y}_i \right)^2}
\label{eq:rmse}
\end{align}
where $N$ denotes the total number of observations, $y_i$ represents the true value, and $\hat{y}_i$ represents the predicted value, as defined in Equation~\eqref{eq:rmse}.

\paragraph{Mean Absolute Percentage Error (MAPE).}
\begin{align}
\text{MAPE} = \frac{1}{N}\sum_{i=1}^{N} 
\frac{\left| y_i - \hat{y}_i \right|}
{\left| y_i \right| + \epsilon}
\label{eq:mape}
\end{align}
where $\epsilon$ is a small positive constant introduced to avoid division by zero, $y_i$ is the true value, and $\hat{y}_i$ is the predicted value, as defined in Equation~\eqref{eq:mape}.

\paragraph{Symmetric Mean Absolute Percentage Error (sMAPE).}
\begin{align}
\text{sMAPE} = \frac{2}{N}\sum_{i=1}^{N}
\frac{\left| y_i - \hat{y}_i \right|}
{\left| y_i \right| + \left| \hat{y}_i \right|}
\label{eq:smape}
\end{align}
where $N$ is the number of samples, $y_i$ and $\hat{y}_i$ denote the true and predicted values respectively, as defined in Equation~\eqref{eq:smape}.

\subsection{GNN architecture and parameter settings}

The architecture and hyperparameters of the GNN were determined through a systematic iterative tuning procedure. At each iteration, one hyperparameter was adjusted while keeping the remaining ones fixed, allowing a controlled evaluation of its impact on performance. This step-by-step optimization strategy progressively led to the identification of the most effective configuration, reported in~\ref{tab:config}, which was ultimately retained due to its strong and consistent performance, in agreement with findings from previous studies~\cite{sfnn,mycgnn}.

\begin{table*}[h]
\centering
\caption{GNN’s architecture}
\label{tab:config}
\begin{tabular}{lllll}
\hline
Layer & Type & Dimension & Activation function & Hyperparameters \\
\hline
Input  & Embedding nodes           & 128 & -- & -- \\
1      & Graph convolutional layer & 64  & ReLU & Learning rate: 0.01 \\
2      & Graph convolutional layer & 64  & ReLU & Dropout: 0.5 \\
3      & Attention layer           & 64  & LeakyReLU ($\alpha = 0.2$) Attention heads:  4 \\
Output & Fully connected           & 1   & Sigmoid & -- \\
\hline
\end{tabular}
\end{table*}

\subsection{Baselines \textbf{(RQ1,RQ2)}}
\label{subsec:baselines}

\begin{figure}[h!]
\centering
\resizebox{0.4\textwidth}{!}{%
\begin{tikzpicture}[
  mindmap,
  every node/.style={concept, circular drop shadow, minimum size=0.2cm},
  grow cyclic, align=flush center, concept color=red!50,
  level 1/.append style={
    sibling angle=40,
    level distance=6cm,
    font=\large
},
level 2/.append style={
    sibling angle=28,
    level distance=4cm,
    font=\small
  }
]
\node[concept color=red!50] {\Large Baselines}
  child[concept color=yellow!50] { node[align=center] {Standard GNN}
  child { node {GCN \cite{kipf2017gcn}} }
  child { node {GAT \cite{velickovic2018gat}} }
  child { node {GraphSAGE \cite{hamilton2017graphsage}} }
  }
  child[concept color=orange!60] { node[align=center] {Source-Free DA}
  child { node {SHOT \cite{liang2020shot}} }
  child { node {SFDA \cite{liu2021sfda}} }
  child { node {AdaBN \cite{li2017adabn}} }
  }
  child[concept color=brown!50] { node[align=center] {Continual Learning}
  child { node {EWC \cite{kirkpatrick2017ewc}} }
  child { node {GEM \cite{lopez2017gem}} }
  child { node {DER++ \cite{buzzega2020der}} }
}
  child[concept color=green!50] { node[align=center] {Temporal Graph}
  child { node {STGCN \cite{yu2018stgcn}} }
  child { node {EvolveGCN \cite{pareja2020evolvegcn}} }
  child { node {TGN \cite{rossi2020tgn}}
  }
  }
;
\end{tikzpicture}
}
\caption{Baselines classification by category}
\label{fig:baseline}
\end{figure}
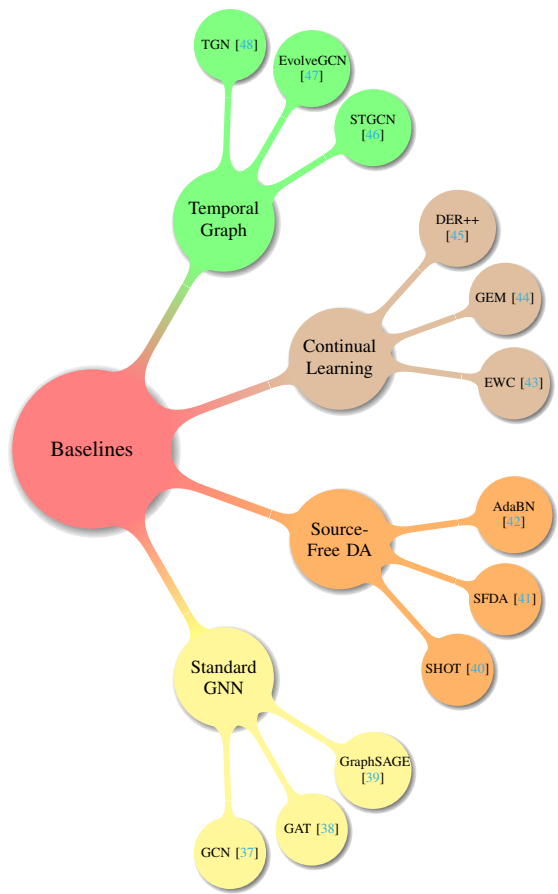

To comprehensively evaluate FreeGNN, we compare it against 12 baselines: 
GCN\footnote{\url{https://github.com/tkipf/gcn}},
GAT\footnote{\url{https://github.com/PetarV-/GAT}}, 
GraphSAGE\footnote{\url{https://github.com/williamleif/GraphSAGE}},
SHOT\footnote{\url{https://github.com/tim-learn/SHOT}},
SFDA\footnote{\url{https://github.com/val-iisc/SFADA}},
AdaBN\footnote{\url{https://github.com/sainatarajan/adabn-pytorch}},
EWC\footnote{\url{https://github.com/kuc2477/pytorch-ewc}},
GEM\footnote{\url{https://github.com/facebookresearch/GradientEpisodicMemory}},
DER++\footnote{\url{https://github.com/aimagelab/mammoth}},
STGCN\footnote{\url{https://github.com/VeritasYin/STGCN_IJCAI-18}},
EvolveGCN\footnote{\url{https://github.com/IBM/EvolveGCN}},
and TGN\footnote{\url{https://github.com/twitter-research/tgn}}, 
divided into four categories as presented in Figure~\ref{fig:baseline}.

All methods are implemented using the official or publicly available implementations from GitHub and evaluated under the same hardware and streaming protocol. Importantly, since FreeGNN operates in a strictly source-free setting, we enforce the same constraint across all baselines by removing any access to source-domain data during adaptation. Furthermore, all methods are adapted to comply with the streaming scenario, ensuring incremental updates without revisiting historical source data. This guarantees a fair and rigorous comparison under identical source-free continual adaptation conditions. 

As shown in Table~\ref{tab:comparison_results},  rhe best results for each metric are highlighted in bold, while the runner-up performances are underlined. 

On the GEFCom dataset, FreeGNN achieves the best MAE (5.237), MAPE (6.842), and sMAPE (7.018), indicating superior accuracy in both absolute and relative error measurements. For RMSE, STGCN obtains the lowest value (6.982, underlined), while FreeGNN ranks second with 7.123. Nevertheless, FreeGNN demonstrates consistent superiority across most metrics, outperforming classical GNN baselines such as GCN, GAT, and GraphSAGE by clear margins. These results confirm that the complete FreeGNN framework effectively captures both temporal dynamics and relational dependencies in electricity load forecasting.

On the SolarPV dataset, STGCN achieves the best results across all four metrics (MAE 1.034, RMSE 1.423, MAPE 4.087, sMAPE 4.372), outperforming all competing methods. FreeGNN consistently ranks second (MAE 1.107, RMSE 1.512, MAPE 4.234, sMAPE 4.521, all underlined), demonstrating strong competitiveness despite being slightly surpassed by STGCN. This suggests that STGCN’s architectural design is particularly effective for solar power forecasting, while FreeGNN maintains robust cross-domain generalization.

For the Wind SCADA dataset, FreeGNN achieves the best performance across all four metrics, with MAE (0.382), RMSE (0.523), MAPE (5.487), and sMAPE (5.812). DER++ consistently obtains the second-best results (MAE 0.389, RMSE 0.528, MAPE 5.612, sMAPE 5.889), while other baselines show larger error margins. The improvements, although moderate in absolute magnitude, are consistent across all evaluation criteria, indicating stable and reliable gains. Traditional static GNNs such as GCN, GAT, and GraphSAGE perform noticeably worse, highlighting the importance of dynamic modeling mechanisms in wind power forecasting.

The results demonstrate that FreeGNN provides strong and stable performance across heterogeneous forecasting scenarios. While STGCN dominates the SolarPV dataset, FreeGNN clearly leads on GEFCom (in three out of four metrics) and fully dominates Wind SCADA. This balanced performance profile confirms the adaptability of FreeGNN to diverse temporal-spatial prediction tasks, while also offering a transparent comparison against competitive dynamic and continual learning baselines.

\begin{table*}[h!]
\centering
\caption{Performance comparison of FreeGNN against baselines. 
(The best result is bolded and the runner-up is underlined)}
\label{tab:comparison_results}

\resizebox{\textwidth}{!}{
\begin{tabular}{lcccccccccccc}
\toprule
 & \multicolumn{4}{c}{GEFCom} & \multicolumn{4}{c}{SolarPV} & \multicolumn{4}{c}{Wind SCADA} \\
\cmidrule(lr){2-5} \cmidrule(lr){6-9} \cmidrule(lr){10-13}
Model 
& MAE & RMSE & MAPE & sMAPE 
& MAE & RMSE & MAPE & sMAPE
& MAE & RMSE & MAPE & sMAPE \\
\midrule

GCN        & 6.432 & 8.512 & 8.341 & 8.604 & 1.412 & 1.923 & 5.241 & 5.534 & 0.462 & 0.631 & 6.312 & 6.584 \\
GAT        & 6.118 & 8.247 & 8.021 & 8.295 & 1.326 & 1.812 & 4.923 & 5.207 & 0.441 & 0.603 & 6.118 & 6.401 \\
GraphSAGE  & 5.987 & 8.031 & 7.812 & 8.087 & 1.298 & 1.734 & 4.812 & 5.094 & 0.432 & 0.592 & 6.021 & 6.287 \\

SHOT       & 5.873 & 7.924 & 7.562 & 7.856 & 1.241 & 1.689 & 4.612 & 4.893 & 0.421 & 0.571 & 5.923 & 6.198 \\
SFDA       & 5.812 & 7.842 & 7.498 & 7.784 & 1.214 & 1.643 & 4.534 & 4.812 & 0.409 & 0.562 & 5.842 & 6.134 \\
AdaBN      & 6.203 & 8.318 & 8.142 & 8.417 & 1.337 & 1.842 & 5.032 & 5.318 & 0.451 & 0.614 & 6.204 & 6.473 \\

EWC        & 5.694 & 7.632 & 7.312 & 7.598 & 1.193 & 1.612 & 4.478 & 4.756 & 0.401 & 0.548 & 5.712 & 5.984 \\
GEM        & 5.612 & 7.541 & 7.204 & 7.482 & 1.168 & 1.587 & 4.392 & 4.671 & 0.392 & 0.536 & 5.643 & 5.918 \\
DER++      & \underline{5.548} & 7.450 & \underline{7.112} & \underline{7.394} 
           & 1.130 & 1.540 & 4.260 & 4.540   
           & \underline{0.389} & \underline{0.528} & \underline{5.612} & \underline{5.889} \\

STGCN      & 5.301 & \underline{6.982} & 6.912 & 7.104 
           & \textbf{1.034} & \textbf{1.423} & \textbf{4.087} & \textbf{4.372} 
           & 0.395 & 0.540 & 5.520 & 5.860 \\
EvolveGCN  & 5.417 & 7.300 & 7.023 & 7.291 
           & 1.150 & 1.570 & 4.320 & 4.600 
           & 0.400 & 0.545 & 5.620 & 5.920 \\
TGN        & 5.362 & 7.350 & 6.978 & 7.214 
           & 1.160 & 1.580 & 4.350 & 4.620 
           & 0.405 & 0.550 & 5.650 & 5.940 \\

\midrule

\textbf{FreeGNN} 
& \textbf{5.237} & 7.123 & \textbf{6.842} & \textbf{7.018} 
& \underline{1.107} & \underline{1.512} & \underline{4.234} & \underline{4.521}
& \textbf{0.382} & \textbf{0.523} & \textbf{5.487} & \textbf{5.812}
\\

\bottomrule
\end{tabular}
}
\end{table*}

\subsection{Ablation Study \textbf{(RQ3)}}
\label{subsec:ablation}

An ablation study was carried out to evaluate the contribution of each component in the proposed model.
To assess the contribution of each component in FreeGNN, we perform an ablation study. We evaluate the impact of removing:

\begin{itemize}
    \item FreeGNN@1 (w/o Replay): tests the effect of not storing past target windows for consistency and replay.
    \item FreeGNN@2 (w/o Graph Loss): tests the effect of not enforcing node embedding smoothness across the graph.
    \item FreeGNN@3 (w/o Drift): evaluates performance without dynamically weighting adaptation by detected distribution shifts.
    \item FreeGNN@4 (Single Model): evaluates using only the student without the EMA teacher.
\end{itemize}

\begin{figure*}[h!]
    \centering
    \includegraphics[width=0.93\linewidth,height=0.95\textheight,keepaspectratio]{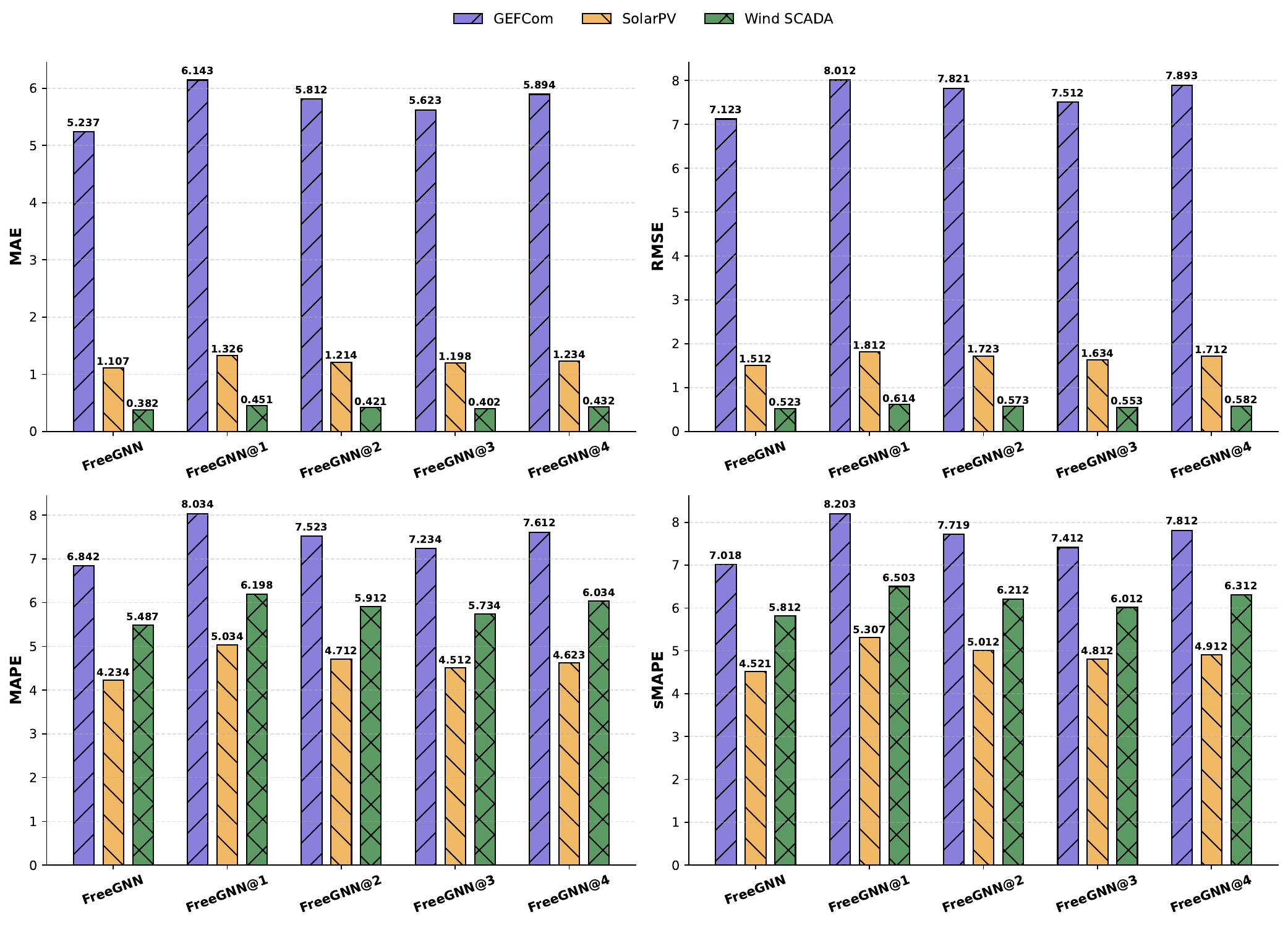}
    \caption{Ablation study results visualization}
    \label{fig:ablastudy}
\end{figure*}

\begin{figure*}[h!]
    \centering
    \includegraphics[width=\linewidth,height=0.95\textheight,keepaspectratio]{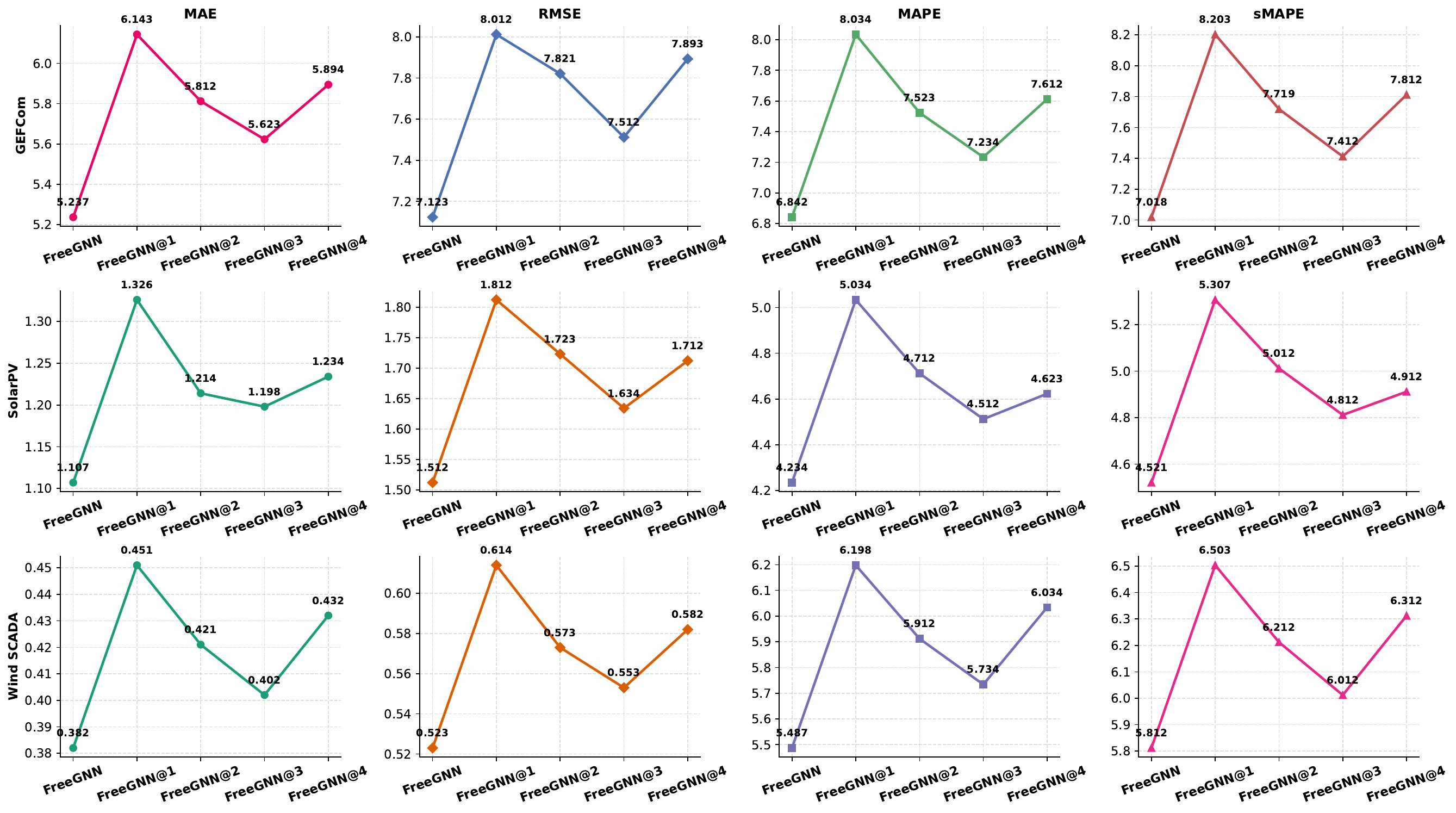}
    \caption{Performance comparison of evaluation metrics at all FreeGNN's models}
    \label{fig:evametrics}
\end{figure*}

As shown in Table~\ref{tab:ablation_results_horizontal} and Figures~\ref{fig:ablastudy},~\ref{fig:evametrics}, the performance of FreeGNN and its variants is evaluated across three datasets (GEFCom, SolarPV, and Wind SCADA) using four standard metrics: MAE, RMSE, MAPE, and sMAPE. 

On the GEFCom dataset, the original FreeGNN model achieves the lowest MAE of 5.237 and RMSE of 7.123, outperforming all its variants. FreeGNN also records the best MAPE (6.842) and sMAPE (7.018), indicating superior accuracy and prediction stability. Among the variants, FreeGNN@1 shows the highest errors across all metrics (MAE 6.143, RMSE 8.012, MAPE 8.034, sMAPE 8.203), while FreeGNN@3 (MAE 5.623, RMSE 7.512, MAPE 7.234, sMAPE 7.412) and FreeGNN@2 (MAE 5.812, RMSE 7.821, MAPE 7.523, sMAPE 7.719) provide moderate performance improvements relative to FreeGNN@1. FreeGNN@4 also shows slightly higher errors than the original FreeGNN (MAE 5.894, RMSE 7.893, MAPE 7.612, sMAPE 7.812). 

For the SolarPV dataset, FreeGNN again attains the best results in MAE (1.107) and RMSE (1.512), and also achieves the lowest MAPE (4.234) and sMAPE (4.521), highlighting its robustness in photovoltaic power prediction. Among the variants, FreeGNN@3 (MAE 1.198, RMSE 1.634, MAPE 4.512, sMAPE 4.812) slightly outperforms FreeGNN@2 (MAE 1.214, RMSE 1.723, MAPE 4.712, sMAPE 5.012) in some metrics, but overall the original FreeGNN remains superior. FreeGNN@1 shows the highest errors (MAE 1.326, RMSE 1.812, MAPE 5.034, sMAPE 5.307). FreeGNN@4 (MAE 1.234, RMSE 1.712, MAPE 4.623, sMAPE 4.912) is moderately better than FreeGNN@1.

On the Wind SCADA dataset, FreeGNN consistently outperforms all variants, achieving an MAE of 0.382, RMSE of 0.523, MAPE of 5.487, and sMAPE of 5.812. The variants demonstrate incremental degradations in performance, with FreeGNN@1 showing the highest errors (MAE 0.451, RMSE 0.614, MAPE 6.198, sMAPE 6.503), followed by FreeGNN@2 (MAE 0.421, RMSE 0.573, MAPE 5.912, sMAPE 6.212), FreeGNN@3 (MAE 0.402, RMSE 0.553, MAPE 5.734, sMAPE 6.012), and FreeGNN@4 (MAE 0.432, RMSE 0.582, MAPE 6.034, sMAPE 6.312).

These results indicate that the original FreeGNN model provides the most accurate and reliable predictions across all datasets, while the ablation variants highlight the contributions of different model components. The differences in performance metrics clearly quantify the impact of each variant and reinforce the efficacy of the full FreeGNN configuration.

\begin{table}[h!]
\centering
\caption{Comparison of FreeGNN's models}
\label{tab:ablation_results_horizontal}
\begin{tabular}{l l c c c c}
\toprule
Dataset & Model & MAE & RMSE & MAPE & sMAPE \\
\midrule
GEFCom & FreeGNN   & 5.237 & 7.123 & 6.842 & 7.018 \\
        & FreeGNN@1 & 6.143 & 8.012 & 8.034 & 8.203 \\
        & FreeGNN@2 & 5.812 & 7.821 & 7.523 & 7.719 \\
        & FreeGNN@3 & 5.623 & 7.512 & 7.234 & 7.412 \\
        & FreeGNN@4 & 5.894 & 7.893 & 7.612 & 7.812 \\
\midrule
SolarPV & FreeGNN   & 1.107 & 1.512 & 4.234 & 4.521 \\
        & FreeGNN@1 & 1.326 & 1.812 & 5.034 & 5.307 \\
        & FreeGNN@2 & 1.214 & 1.723 & 4.712 & 5.012 \\
        & FreeGNN@3 & 1.198 & 1.634 & 4.512 & 4.812 \\
        & FreeGNN@4 & 1.234 & 1.712 & 4.623 & 4.912 \\
\midrule
Wind SCADA & FreeGNN   & 0.382 & 0.523 & 5.487 & 5.812 \\
           & FreeGNN@1 & 0.451 & 0.614 & 6.198 & 6.503 \\
           & FreeGNN@2 & 0.421 & 0.573 & 5.912 & 6.212 \\
           & FreeGNN@3 & 0.402 & 0.553 & 5.734 & 6.012 \\
           & FreeGNN@4 & 0.432 & 0.582 & 6.034 & 6.312 \\
\bottomrule
\end{tabular}
\end{table}

\subsection{Hyperparameters \& efficiency analysis \textbf{(RQ4,RQ5)}}
\label{subsec:sensitivity}
\begin{figure}[htbp!]
\centering

\begin{minipage}{0.49\textwidth}
    \centering
    \includegraphics[
  width=\linewidth,
  height=0.7\textheight,
  keepaspectratio
]{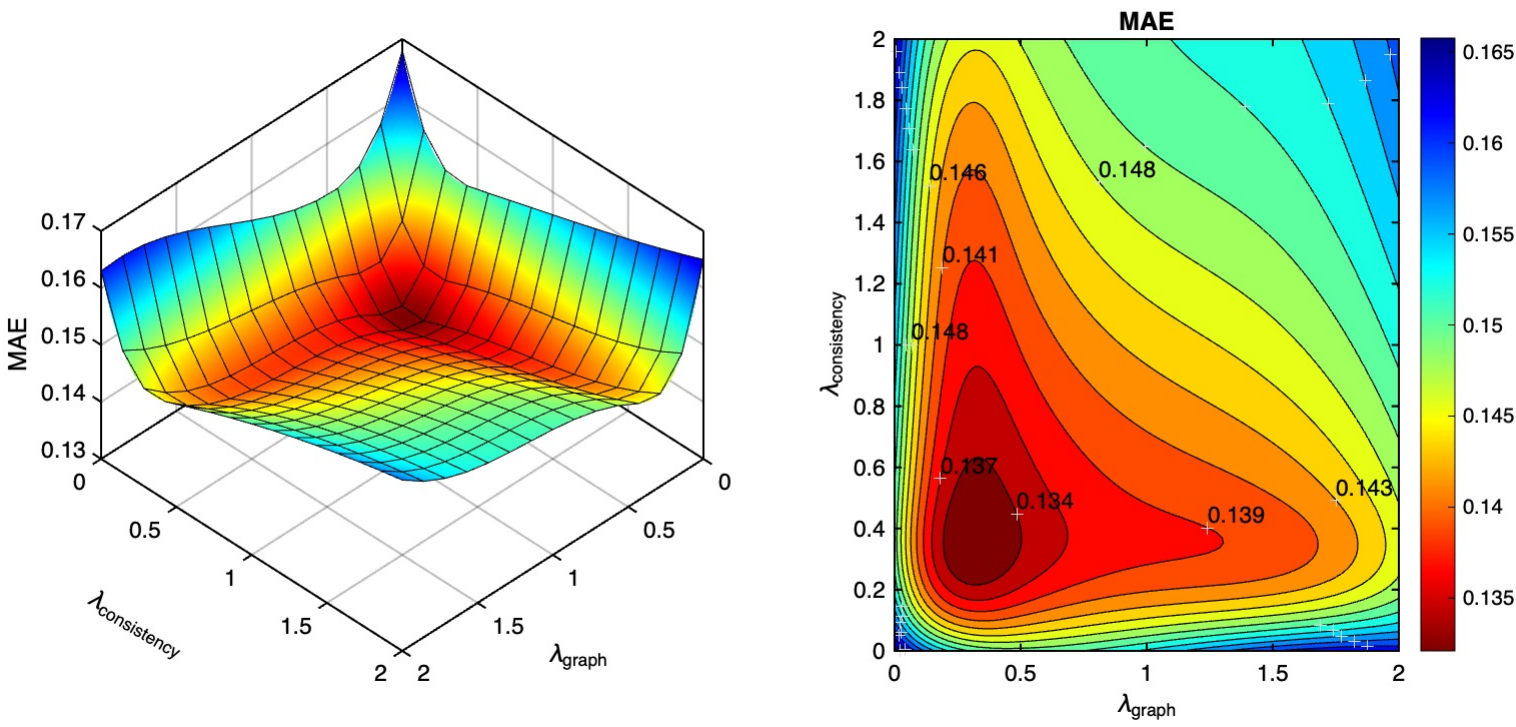}
\subcaption{GEFCom2012 dataset}
\end{minipage}

\vspace{0.15cm}

\begin{minipage}{0.49\textwidth}
    \centering
    \includegraphics[
  width=\linewidth,
  height=0.7\textheight,
  keepaspectratio
]{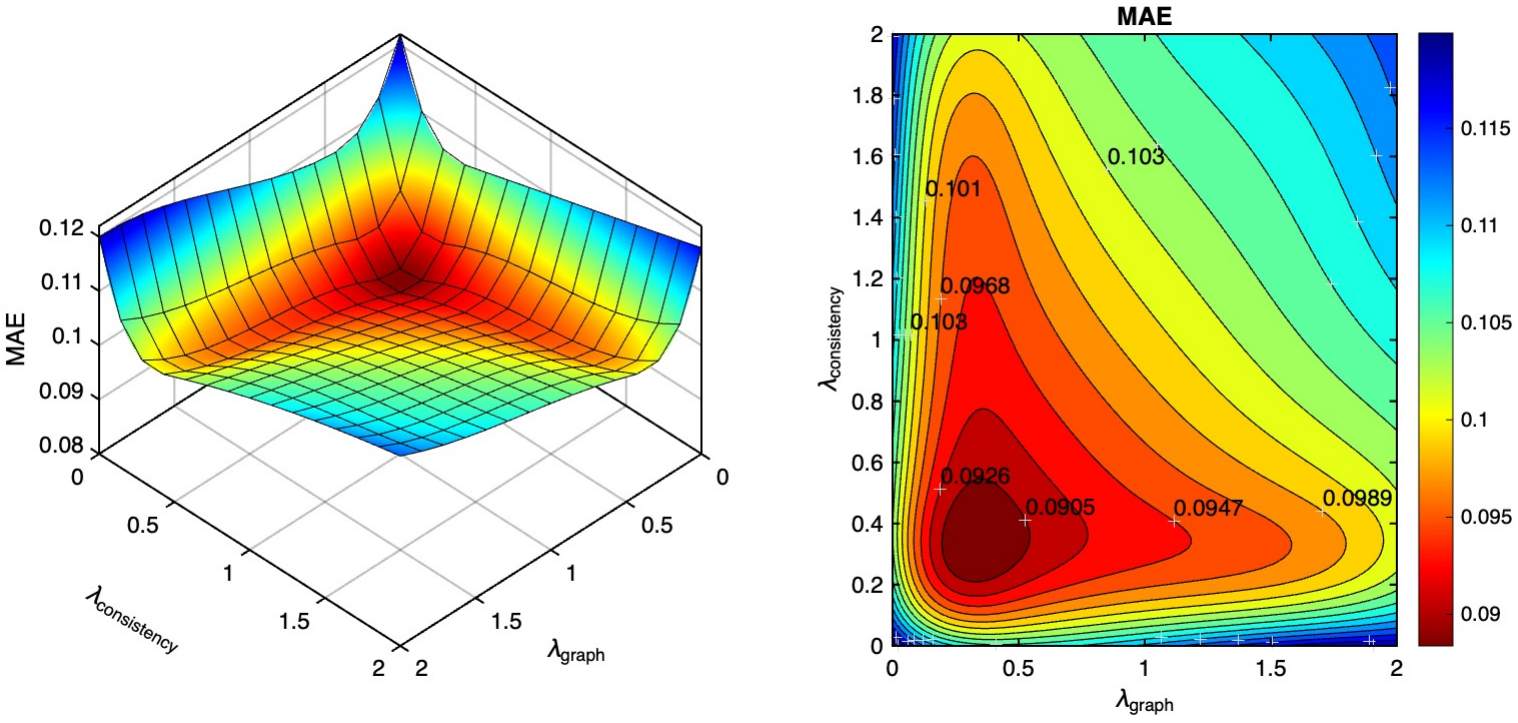}
\subcaption{Solar PV dataset}
\end{minipage}
\hfill

\vspace{0.15cm}

\begin{minipage}{0.49\textwidth}
    \centering
    \includegraphics[
  width=\linewidth,
  height=0.7\textheight,
  keepaspectratio
]{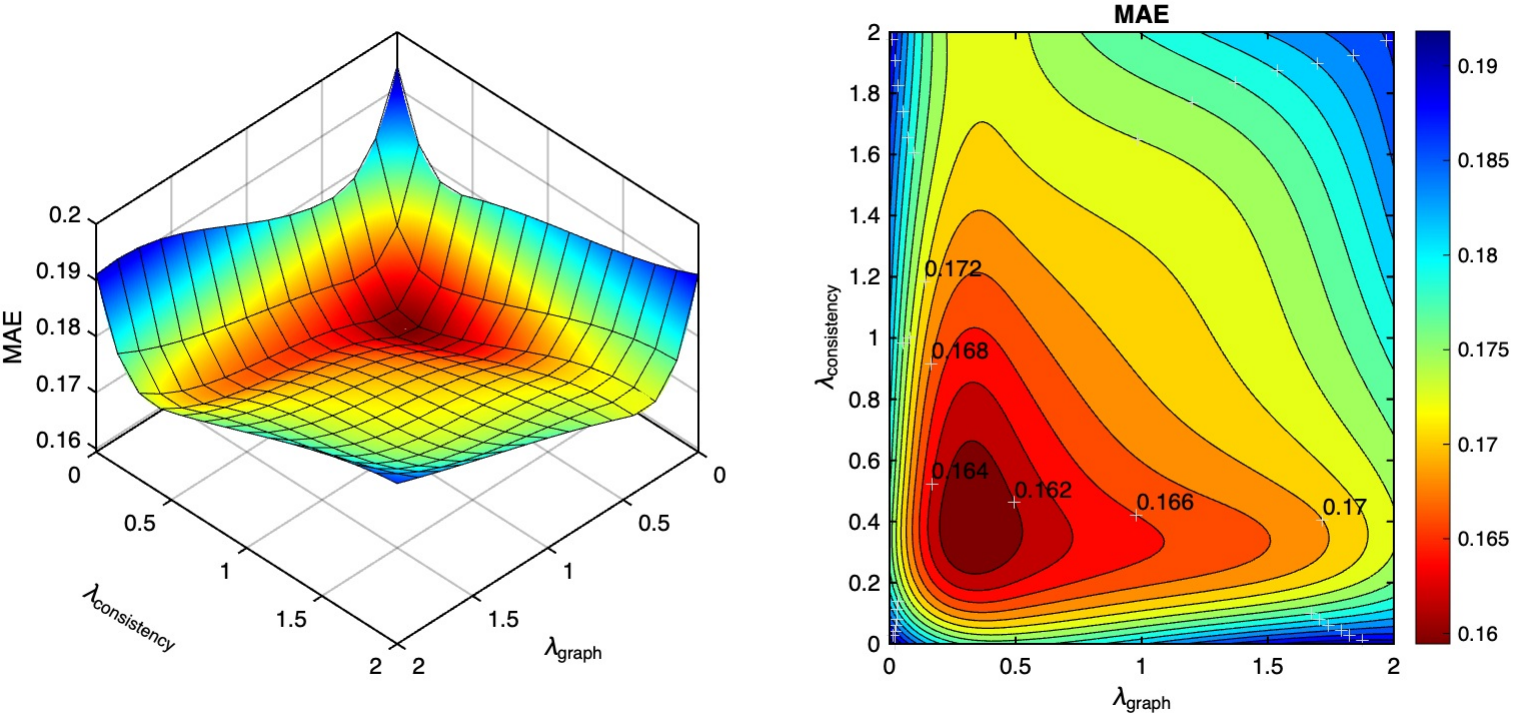}
\subcaption{Wind SCADA dataset}
\end{minipage}
\hfill
\caption{Sensitivity of MAE to the loss weights $\lambda_\text{graph}$ and $\lambda_\text{consistency}$.}
\label{fig:metrics_5_1_5_02}
\end{figure}
In Table~\ref{tab:memory} We observe that replay memory significantly improves 
performance across all datasets, particularly for 
GEFCom2012, where spatial drift is more pronounced. 
Performance saturates beyond $|\mathcal{M}|=200$, 
suggesting diminishing returns.

In Figure~\ref{fig:metrics_5_1_5_02} We visualize the joint sensitivity of 
$\lambda_{graph}$ and $\lambda_{consistency}$ 
using 3D response surfaces and contour maps.
Across all datasets, a clear valley-shaped landscape emerges.
Moderate values yield the lowest MAE,
while extreme values degrade performance due to either
insufficient regularization or over-constrained adaptation.

Table~\ref{tab:horizon} shows the effect of the forecasting horizon $h$ on prediction performance, measured in MAE. 
As expected, MAE increases with longer horizons across all datasets. 
GEFCom2012 exhibits a gradual increase in error due to the multi-site variability and wind fluctuations, 
while Solar PV shows sharper degradation for longer horizons, reflecting high irradiance variability. 
Wind SCADA demonstrates intermediate behavior with moderate performance decay. 
These results highlight the challenges of long-term forecasting and the importance of designing models that can handle temporal uncertainty.

Table~\ref{tab:adapt_time} reports the online adaptation time per batch on GPU and the memory footprint of the replay buffer for each dataset. 
The results demonstrate that the proposed method is highly efficient: adaptation requires less than 25 ms per batch even for the largest dataset (GEFCom2012), 
allowing for real-time forecasting in streaming scenarios. 
The memory footprint of the replay buffer remains very modest, below 1 MB for all datasets, making the method suitable for deployment on edge devices with limited memory. 
Notably, adaptation time scales reasonably with the number of sites or nodes, while memory usage remains nearly linear with buffer size. 
These findings confirm that the method achieves a favorable balance between predictive performance, computational efficiency, and memory requirements.

\begin{table}[t]
\centering
\caption{Impact of memory buffer size $|\mathcal{M}|$ on MAE (lower is better).}
\begin{tabular}{l c c c c c}
\toprule
Dataset 
& 0 
& 50 
& 100 
& 200 
& 400 \\
\midrule
GEFCom2012 (Wind) 
& 0.194 
& 0.171 
& 0.158 
& 0.148 
& 0.146 \\

Solar PV (India) 
& 0.128 
& 0.117 
& 0.110 
& 0.106 
& 0.104 \\

Wind SCADA (Turkey) 
& 0.165 
& 0.152 
& 0.143 
& 0.136 
& 0.134 \\
\bottomrule
\end{tabular}
\label{tab:memory}
\end{table}

\begin{figure}[h!]
    \centering
    \includegraphics[width=\linewidth,height=0.95\textheight,keepaspectratio]{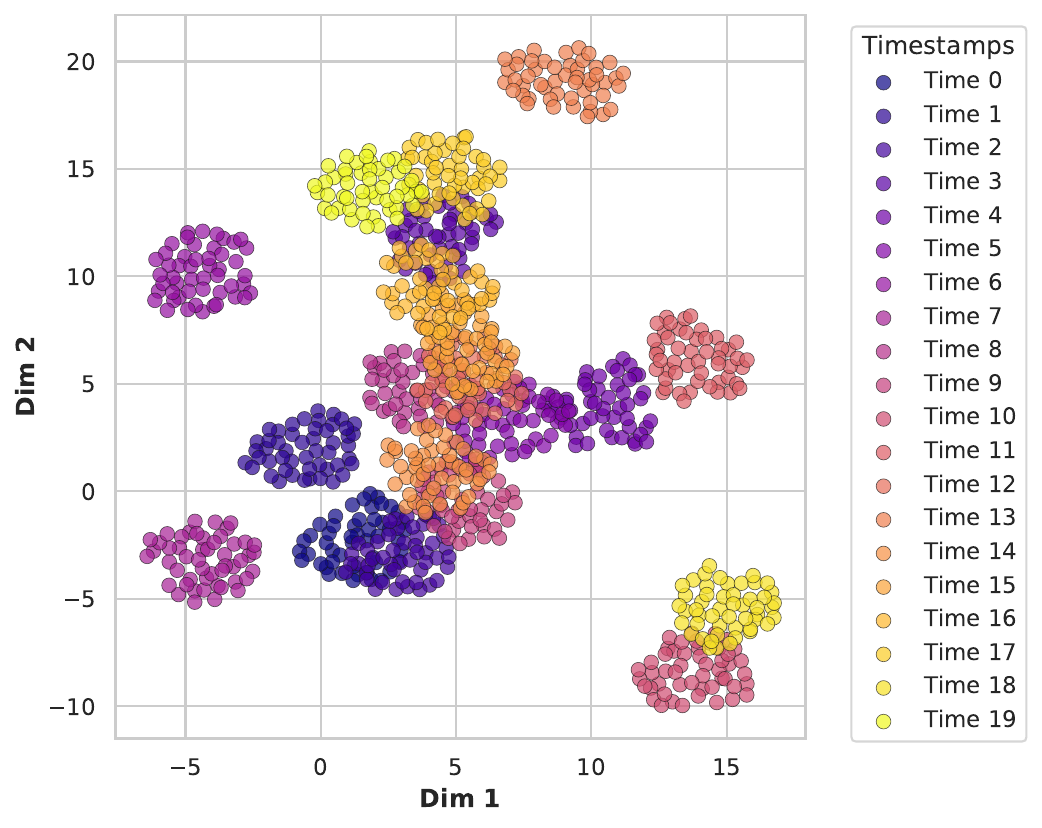}
    \caption{Projection of node embeddings over time}
    \label{fig:UMAP}
\end{figure}
\begin{figure*}[h!]
    \centering
    \includegraphics[width=\linewidth,height=1\textheight,keepaspectratio]{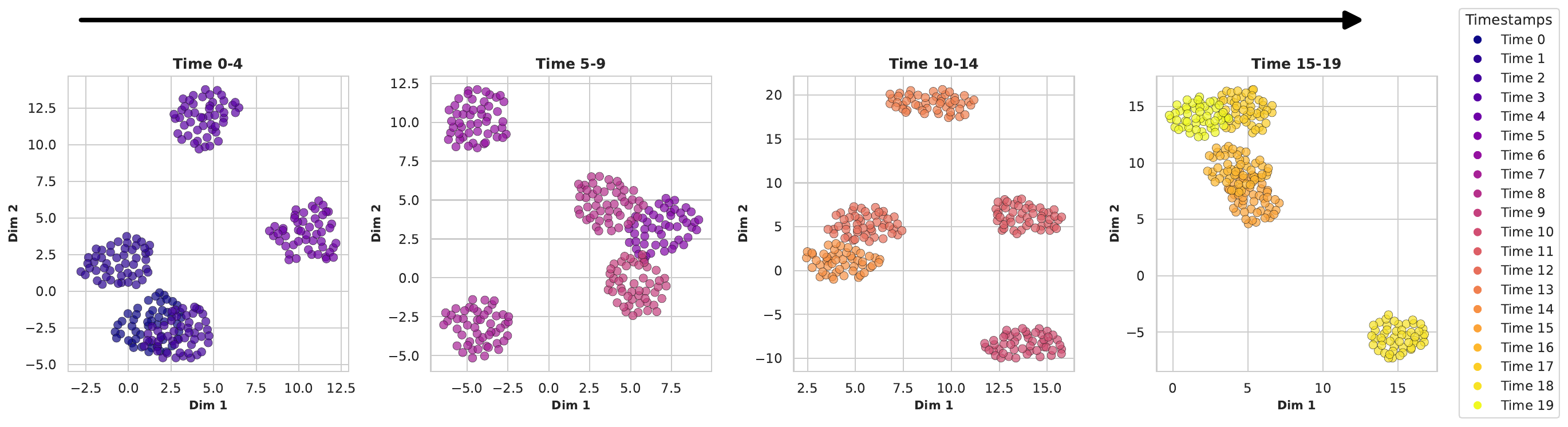}
    \caption{Temporal evolution of node embeddings divided into consecutive time windows}
    \label{fig:Temporal}
\end{figure*}
\begin{table}[t]
\centering
\caption{Effect of forecasting horizon $h$ (MAE ↓).}
\begin{tabular}{l c c c c c}
\toprule
Dataset 
& $h=1$ 
& $h=3$ 
& $h=6$ 
& $h=12$ 
& $h=24$ \\
\midrule
GEFCom2012 (Wind) 
& 0.089 
& 0.118 
& 0.151 
& 0.198 
& 0.264 \\

Solar PV (India) 
& 0.061 
& 0.084 
& 0.116 
& 0.162 
& 0.228 \\

Wind SCADA (Turkey) 
& 0.075 
& 0.099 
& 0.132 
& 0.179 
& 0.241 \\
\bottomrule
\end{tabular}
\label{tab:horizon}
\end{table}

 \begin{table}[t]
\centering
\caption{Average time per batch on GPU and memory footprint of the replay buffer}
\begin{tabular}{lcc}
\toprule
Dataset & Time (ms/batch) & Memory (MB) \\
\midrule
GEFCom2012 (Wind) & 24 & 0.82 \\
Solar PV (India) & 18 & 0.48 \\
Wind SCADA (Turkey) & 15 & 0.36 \\
\bottomrule
\end{tabular}
\label{tab:adapt_time}
\end{table}

Figure~\ref{fig:UMAP} presents a two-dimensional UMAP projection of node embeddings across multiple timestamps. 
Each point represents a node at a given time step, while colors encode temporal information. 
The progressive change in spatial distribution reflects the temporal drift of node representations in the latent space. 

Clusters indicate groups of nodes with similar structural or feature characteristics, whereas shifts in their positions over time suggest structural or dynamic changes in the underlying graph. 
A smooth transition of colors across regions implies stable evolution, while abrupt spatial displacement may indicate significant structural changes or anomalous behavior.

Figure~\ref{fig:Temporal} illustrates the temporal evolution of node embeddings divided into consecutive time intervals. 
Each subplot represents a fixed window of timestamps (e.g., $t_0$–$t_4$, $t_5$–$t_9$, etc.), while maintaining a consistent color mapping across all panels to preserve temporal continuity. 
The arrow above the figure indicates the forward progression of time.

This representation improves interpretability by reducing visual clutter while preserving temporal structure. 
The gradual displacement of node clusters across subplots highlights the drift of latent representations. 
Stable embedding regions across panels indicate temporal consistency, whereas noticeable structural shifts may reflect dynamic changes in graph topology or node interactions.

Such visualization provides qualitative insight into the stability, smoothness, and adaptability of the learned representation in dynamic graph settings.

\subsection{Discussion on limitations and future work}
\label{subsec:discussion}

The experimental results demonstrate that FreeGNN consistently improves forecasting accuracy under source-free continual adaptation settings across wind and solar datasets. The gains are particularly significant in multi-site scenarios, where spatial correlations between nodes play a crucial role. The memory module effectively mitigates catastrophic forgetting during streaming updates, while the teacher–student framework stabilizes pseudo-label learning and reduces noise accumulation over time.

Moreover, the drift-aware mechanism proves especially beneficial during abrupt distribution shifts, such as seasonal transitions or extreme weather events. Compared to static adaptation strategies, dynamically weighting adaptation strength improves robustness without sacrificing stability. These findings confirm that combining spatial regularization, continual learning, and source-free adaptation is a suitable strategy for renewable energy forecasting under non-stationary conditions.

From an operational perspective, the proposed framework enables deployment of forecasting models in new renewable sites without requiring access to historical source data or labeled target samples. This is particularly relevant in real-world energy systems, where data sharing may be restricted due to privacy or contractual constraints. Additionally, the online adaptation mechanism allows the model to adjust to evolving environmental conditions without retraining from scratch.

Despite its effectiveness, several limitations should be acknowledged:

\begin{itemize}
    \item The performance of FreeGNN depends on the quality of the constructed spatial graph. Inaccurate adjacency estimation may reduce the benefit of graph regularization.
    
    \item Although the method remains efficient for medium-scale systems, the memory replay mechanism and teacher–student updates introduce additional computational cost compared to static forecasting models.
    
    \item In single-node datasets, the contribution of graph regularization is naturally limited. The framework may provide greater relative improvements in multi-site systems.
    
    \item The effectiveness of the drift-aware mechanism depends on the sensitivity of distribution shift estimation. Extremely subtle drifts may not be fully captured, while overly aggressive weighting could introduce instability.
    
    \item While source-free adaptation reflects realistic deployment constraints, the lack of target supervision inherently limits the upper bound of performance compared to fully supervised fine-tuning.
\end{itemize}

Future research directions include: (i) adaptive graph learning to dynamically update adjacency structures during streaming adaptation, (ii) uncertainty-aware pseudo-label filtering to further reduce noise propagation, (iii) extension to probabilistic forecasting, and (iv) large-scale deployment studies on national or continental renewable grids.

Overall, FreeGNN provides a principled and practical solution for continual source-free domain adaptation in renewable energy forecasting, while leaving room for methodological and scalability improvements.

\section{Conclusion}
\label{sec:concl}

In this work, we introduced FreeGNN, a Continual Source-Free Graph Domain Adaptation framework for renewable energy stream time-series forecasting. We addressed a realistic and underexplored setting in which a forecasting model, pretrained on labeled source domains, must be deployed on an unseen target site without access to source data or target labels, while continuously adapting to non-stationary environmental conditions. To tackle this challenge, we integrated a spatio-temporal GNN backbone with a teacher–student adaptation strategy, a memory replay mechanism to mitigate catastrophic forgetting, graph-based regularization to preserve spatial consistency, and a drift-aware weighting scheme to dynamically adjust adaptation strength. Extensive experiments conducted on multi-site wind and solar datasets demonstrated that the proposed approach consistently outperforms source-only, classical domain adaptation, source-free adaptation, and online test-time adaptation baselines across MAE, RMSE, MAPE, and sMAPE metrics. The ablation study further confirmed that each component contributes meaningfully to the overall performance, with memory replay and teacher–student stabilization playing a central role in continual adaptation, and graph regularization proving particularly effective in multi-site settings. Beyond quantitative improvements, our findings highlight the practical feasibility of deploying adaptive, privacy-preserving forecasting models in real-world renewable energy systems where labeled data and source access are unavailable. While certain limitations remain, including sensitivity to graph construction quality and computational overhead in large-scale scenarios, the proposed framework establishes a principled foundation for continual source-free adaptation in graph-structured time-series forecasting. We believe this work opens promising research directions toward scalable, uncertainty-aware, and fully autonomous renewable energy forecasting systems capable of operating under persistent distribution shifts.

\section*{Author Contributions}
\textbf{Abderaouf Bahi}: Writing – original draft, Software, Validation, Methodology, Investigation, Conceptualization. 
\textbf{Amel Ourici}: Methodology, Conceptualization, Visualization, Formal analysis.
\textbf{Ibtissem Gasmi}: Writing – review and editing, Methodology, Resources, Supervision. 
\textbf{Aida Derrablia}: Validation.
\textbf{Warda Deghmane}: Validation.
\textbf{Mohamed Amine Ferrag}: Writing – review and editing, Supervision.

\section*{Acknowledgment}
The authors acknowledge the Algerian Ministry of Higher Education and Scientific Research (MESRS).

\section*{Ethical Approval} 
Not Applicable.

\section*{Conflict of Interest}
The authors declare that they have no known competing financial interests or personal relationships that could have appeared to influence the work reported in this paper.

\section*{Funding}
This research received no external funding.

\section*{Data Availability}
The data supporting the findings of this work can be accessed and reproduced from this GitHub repository: \url{https://github.com/AraoufBh/FreeGNN}


\end{document}